# Heterogeneous Resource Allocation for Ensuring End-to-End Quality of Service in Multi-hop Integrated Access and Backhaul Networks

Shuaifeng Zhang

School of Computer Science and Engineering, Central South University, Changsha, 410083 China

*Abstract*—Faced with increasing network traffic demands, cell dense deployment is one of significant means to utilize spectrum resources efficiently to improve network capacity. Multi-hop integrated access and backhaul (IAB) architectures have emerged as a cost-effective solution for network densification. Meanwhile, dynamic time division duplex (D-TDD) is a promising solution to adapt to highly dynamic scenarios with asymmetric uplink and downlink traffic. Thus, dynamic resource allocation between backhaul and access links and high spectral efficiency under ensuring reliable transmission are two key objectives of IAB research. However, due to huge solution space, there are some challenges in multi-hop IAB with D-TDD if only an integrated optimization problem (IOP) is considered. To handle these challenges, we decompose the IOP into sub-problems to reduce the solution space. To tackle these sub-problems, we formulate them separately as the non-cooperative games and design the corresponding utility functions to guarantee the existence of Nash equilibrium solutions. Also, to achieve the system-wide solution, we propose a single-leader heterogeneous multi-follower Stackelberg-game-based resource allocation scheme, which can combine the solving results of all the sub-problems to get the IOP approximate solution. Simulation results show that the proposed scheme can improve throughput performance while meeting spectrum energy efficiency constraints.

## 1. INTRODUCTION

In order to meet the demand for high-data traffic in future mobile communication systems, a lot of techniques have been explored. One widely recognized approach is to extend the spectrum range from sub-6GHz to millimeter wave (mmWave) even terahertz (THz) spectrum to provide high-data rate. At the same time, the improvement of spectrum utilization is also an indispensable means. The flexible and dynamic duplex operation can effectively improve spectrum utility [1]. The long-term evolution (LTE) systems (i.e., Fourth generation mobile communication system, 4G) are using the two duplex modes: frequency division duplex (FDD) and time division duplex (TDD). FDD mode requires the uplink (UL) and downlink (DL) to operate in different frequency bands, while TDD mode allows them to operate in different time slots of the same frequency band. Compared with TDD mode, FDD mode occupies more spectrum resources. Moreover, some scattered spectrum resources cannot be used in FDD mode, resulting in spectrum waste.

Dynamic Time Division Duplex (D-TDD) is a promising solution to address newly emerging 5G and 6G services characterized by asymmetric and dynamic uplink (UL) and downlink (DL) traffic demands. For example, in urban areas with heavy traffic, the camera mainly uploads data through UL, while in remote urban suburbs, the map software mainly downloads the map information through DL. However, the flexibility of traffic configuration of D-TDD introduces additional inter-cell interference, which largely deteriorates network capacity.

The 3GPP LTE and LTE-Advanced (LTE-A) standards have provided the relevant specifications for the SBSs with radio backhauling capabilities. Although Release 17 of 5G-NR [2] defines the interfaces, architectures, it is the work of operators that the actual network configuration and resource allocation. Also, the study item on IAB [3] expects that the more advanced and flexible solutions, including wireless multi-hop communications and resource dynamic multiplexing, should be supported. There is general consensus about IAB's ability to reduce network deployment costs, but it is still a challenging problem how to design an efficient IAB network with high-performance. Furthermore, it is also an open research challenge how to improve D-TDD-based resource utilization in IAB network architecture.

The existing relevant works solely focused on either D-TDD-based problems or IAB-based problems. However, the work in [4] focused on heuristic implementations of S-TDD and D-TDD for access links with synchronized or unsynchronized access-backhaul time splits. Although it involves all the types of mutual interferences in D-TDD-based resource allocation of IAB networks, it does not involve wireless multi-hop backhaul communications, which limits the cellular coverage based on wireless backhaul. In [4], each time frame with fixed length is first divided into the access fraction and the backhaul fraction, and then each fraction is divided into the DL duration and the UL duration respectively. Thus, each time frame requires three parameters to represent the frame division, which complicates the calculation of various co-channel interference durations.

Although the authors in [4] claimed that their work is the first comprehensive study of UL-DL signal to interference plus noise ratio (SINR) distribution and mean rates in D-TDD enabled mmWave cellular networks, they did not consider end-to-end quality of service (QoS) because there is no wireless multi-hop backhaul path in the scenario they considered. The inadequacy of the existing works motivates us to work on this paper, and the main contributions are listed as follows.

1) We model the heterogeneous resource allocation as an integrated optimization problem (IOP) for ensuring end-to-end QoS in multi-hop IAB networks. The significant difference from the existing works is that we consider all the mutual interferences and power adjustments in multi-hop IAB networks.

2) We decompose the IOP into sub-problems to reduce the solution space since the IOP involves radio frame configurations (RFCs), non-unified access and backhaul transmission duration division, and discrete power allocation of all the ULs and DLs. Thus, it facilitates the distribution of huge solution burden to several nodes for collaborative execution.

3) To tackle all the sub-problems with lower computational complexity, we formulate them as non-cooperative games, separately. Also, to achieve the system-wide solution, we propose a single-leader heterogeneous-multi-follower Stackelberg-game-based resource allocation (SL-HMF-RA) scheme, which can combine the solving results of all the sub-problems to get the IOP approximate solution.

4) The SL-HMF-RA scheme can reasonably allocate heterogeneous resources to optimize the network throughput while meeting the end-to-end QoS constraints. Extensive simulations are conducted to verify performance of the proposed scheme and the comparison schemes, which shows the advantages of our scheme over the comparison schemes.

The remainders of this paper are organized as follows. In Section 2, we review the related works in terms of D-TDD-based and IAB-based problems. The system model, including network description, signal propagation model, power discretization and frame structure design, estimation models for interference and throughput, is described in Section 3. The problem statement, including relay SBS selection and user association, discrete power control, and non-unified access and backhaul transmission duration allocation under different UL/DL traffic ratios, is described in Section 4. The problem solving scheme is presented in Section 5. Section 6 evaluates the simulation results. Finally, we conclude this paper in Section 7.

## 2. RELATED WORK

### 1) D-TDD-based works

The time division long term evolution (TD-LTE) provided a special TDD frame structure to adapt to the asymmetric DL and UL traffic demand. Thereafter, the LTE release (Rel)-8 defined the UL/DL subframe configuration of TDD as the ratio of the number of UL subframes and the number of DL subframes. A total of seven UL/DL subframe configurations were defined in a frame, and each frame length is 10 ms, which consists of 10 successive subframes with 1ms. Each subframe belongs to one of three types (i.e., subframe U, D and S). The LTE-A Rel-12 involved dynamically changing configurations if traffic conditions change [5].

5G new radio (NR) supports a more adaptive frame structure, which is possible to configure both DL and UL symbols within one slot [6], and thus CLI can occur more frequently at the symbol level. Therefore, there is an increasingly urgent need to address the CLI problem. When the dense deployment of small cells is adopted to handle the wireless traffic explosion in 5G, the access nodes and links per unit area are densified [7].

In view of the additional inter-cell interference resulting from the flexibility of D-TDD technology in terms of traffic configuration, the authors in [8] developed a distributed deep reinforcement learning (DRL) for each small cell to learn the selection of RFCs to mitigate the negative impact of CLI. The work in [9] proposes a channel parameter estimation based polynomial CLI canceller and two machine learning (ML) based CLI cancellers to better mitigate the impact of CLI. The authors model the dynamic TDD problem in 5G NR as a linear programming problem [10]. Then, they design Multi-Agent Deep Reinforcement Learning to distribute time slots between the UL and the DL and mitigate the CLI between neighboring cells as the 3GPP standard neither specifies algorithms or solutions to derive the TDD configuration nor solves the cross-link interference.

In [11], the authors propose a Machine Learning (ML)-based solution relaying on Deep Reinforcement Learning (DRL) to allow the base station (or gNB) to self-adapt to the traffic pattern of the cell by periodically adapting the number of slots dedicated to UL and DL. The study in [12] investigates a spatial deep learning-based D-TDD scheme for 6G hotspot networks. The proposed D-TDD scheme improves average rate when it is compared to the state-of-the-art centralized D-TDD scheme [13]. In this article [14], the authors propose an innovative system architecture where a full-duplex unmanned aerial vehicle (UAV) serves as a base station (BS) to enhance the performance of D-TDD networks composed of multiple full-duplex ground users and provides essential mobility and duplexing flexibility.

### 2) IAB-based works

In recent years, there have emerged many works on resource management for different types of networks[15],[16], [17],[18],[19]. IAB networks, as a form of 5G network deployment, have also been extensively studied in terms of resource allocation[20],[21],[22].

The importance of the IAB framework as a cost-effective alternative to the wired backhaul has been recognized by the 3GPP. Indeed, it has completed a Study Item for 3GPP NR Release 16 [3], which investigates architectures, radio protocols, and physical layer aspects for sharing radio resources between access and backhaul links.

In [23], a backhauling topology design framework is developed for an IAB network to generate a DAG that supports the traffic between UEs and the IAB-donor with the highest probability. And the authors design a Bayesian DAG generation algorithm to solve the backhauling topology design problem. In [24], a distributed joint flow control and resource allocation algorithm is proposed for IAB networks, called the Dynamic Slot Reservation (DSR) algorithm, which is completely distributed and does not require finding a network-wide, maximum weight ,thus messages to traverse the network is not required.

In paper [25], the authors investigate the availability of IAB architecture in dynamic aerial-terrestrial networks in terms of coverage probability (CP) and further explore the feasible region of IAB to promote aerial-terrestrial coverage enhancement.The article [26] presents the design principle and validation of a practical SI cancellation (SIC) technique in the case of high transmission power class in mmW-based IAB networks. The authors pay attention to the network's reliability, propose Safehaul, a risk-averse learning-based solution for IAB mmWave networks[27]. However, based on the above work on IAB network, we see that their problem scenes did not include the end-to-end QoS assurance in terms of wireless multi-hop backhaul path with D-TDD.

## 3. SYSTEM MODEL

### 3.1 Network Description

We consider a time-varying UL and DL transmission scenario in an IAB network with D-TDD, where a macro base station (MBS) is deployed at the center of the scenario with a set of $M$ SBSs. The MBS is represented as $m_0$ and the SBSs are represented as the set $\mathcal{M}^- = \{m_1, \ldots, m_m, \ldots, m_M\}$, so the set $\mathcal{M} = \mathcal{M}^- \cup \{m_0\}$ can represent all the BSs including the MBS and SBSs, where $|\mathcal{M}| = M + 1$. All the UEs are randomly distributed in the considered network coverage. We assume that all the BSs have three types of radio interfaces (i.e., sub-6 GHz, mmWave, and THz) and also suppose that each UE has at least one of the above types of radio interfaces.

We assume that each BS is equipped with $3K$ RF chains, which can establish at most $K$ concurrent connections in sub-6 GHz, mmWave, and THz bands, respectively. For each type of radio interface, each UE has only one RF chain and thus the number of possible RF chains ranges from 1 to 3. In the considered network coverage, all the UEs can communicate directly with all the BSs in sub-6 GHz band, and all the SBSs can communicate directly with the MBS in sub-6 GHz band. However, in mmWave and THz bands, not all nodes (including BSs and UEs) can communicate directly with each other. That is, some nodes need to be connected by one or several relay nodes in mmWave and THz bands.

The MBS has a fiber backhaul link and thus acts an IAB donor, while all the SBS only adopt wireless backhaul links and thus act as IAB nodes. The mmWave band has a longer direct communication range than the THz band and a higher transmission capacity than the sub-6 GHz band, so it is more suitable to be used as wireless communication medium between BSs. Each SBS should assume relay obligations to maintain IAB network performance. If a SBS does not act as a relay role due to its unfavorable location, it is called a non-relay SBS and thus forbidden to use the THz band so as not to put more pressure on the backhaul transmission. For a SBS acting as a relay role (i.e., a relay SBS), the THz band is allowed to guarantee the transmission performance of its own UEs on the premise of ensuring the completion of relay tasks.

In this paper, we mainly consider using the sub-6 GHz band for the exchange of control information, though it is suitable for the transmission of short data streams or low volume data. For the transmission of long data streams or high-volume data, we mainly consider using the mmWave band but do not ruled out occasional use of THz band. However, unless the UEs can be directly connected to the MBS through mmWave or THz interfaces, the matching problem of transmission capacity between each SBS's access and backhaul links must be considered. We focus on long-stream performance between stationary nodes in this paper. When it is initiated by a UE, long stream transmission occurs in DL scenarios, such as video data requests to remote servers. However, long stream transmission also occurs in UL scenarios, for example, uploading a large volume of data to a remote server.

In addition, when users use Web browsing mode to continuously watch videos, many DL streams separated by the corresponding short UL requesting packets actually have the similar resource demand to a long video stream. Cooperative multiplayer online games also have the resource requirements similar to a long data stream. Although there are different UL/DL traffic ratios in the above applications, a single SBS only accepts the concurrent requests with the similar UL/DL traffic ratio to facilitate the resource allocation of its access and backhaul links. In addition, a SBS accepts UEs' transmission requests according to a fair scheduling policy. Since there are multi-hop communication requirements in some wireless backhaul transmissions, each SBS must report relevant information to the MBS to determine which SBSes need to undertake the relay tasks.

How to determine relays in a fairer way is not covered in this paper. However, we assume that the MBS will plan at most three types of relay cases as shown in Fig. 1, where Case 1 has the lowest requirement on spectrum resources while Case 3 has the highest one. We assume that a set of spectrum resource blocks can be reused by all the BSs in the considered network coverage. In Case 1, at most $K$ UEs can share the same spectrum resource block through spatial separation of directed antennas of the SBS they are associated with, and also the same spectrum resource block can be reused by the backhaul link of the same SBS in time division mode. In Case 2 and Case 3, the relay SBSs must ensure sufficient spectrum resource blocks to complete the relay tasks before they can consider the data transmission of their associated UEs.

To facilitate the matching of transmission capabilities of links on a transmission path, each relay SBS must carry out data transmission in strict accordance with the access and backhaul time divided by the non-relay SBS in the same transmission path, and then achieves the matching target by increasing spectrum resource blocks on demand. The access segment and backhaul segment of any non-relay SBS are multiplexed in the same spectrum resource block via time division mode, so the backhaul segment is usually allocated more time-domain resources to balance the access and backhaul capacity due to the many-to-one relationship between the access links and the backhaul link of this SBS.

Take UL transmission as an example, each relay SBS in Case 2 only transmits during the specified access time, so more spectrum resource blocks are needed to ensure that its forwarding capacity is not lower than its receiving capacity. Similarly, in Case 3, the relay SBS adjacent to the non-relay SBS in the same transmission path only transmits during the specified access time, though the other relay SBS must transmit during the specified backhaul time. Due to more access traffic aggregated step by step, more spectrum resource blocks may be needed in Case 3.

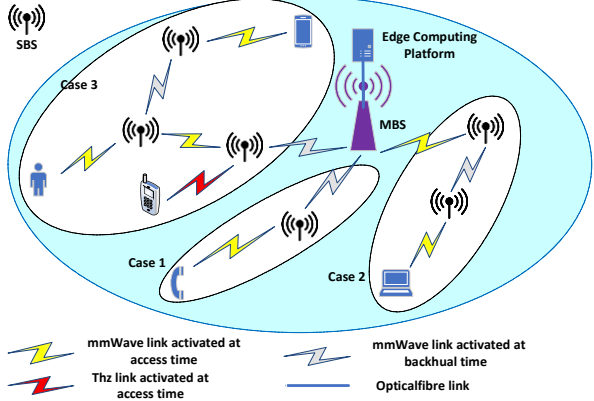

Fig. 1. Illustration of heterogeneous resource allocation in IAB networks

As mentioned above, the access time is shorter than the backhaul time, so any relay SBS that uses the access time to receive or send its associated UEs' data may not meet the QoS demand of its associated UEs. To address this problem, it should preferentially serve the UEs with THz interfaces, where higher data rate can reduce or avoid the loss of transmission performance due to shorter transmission time. In addition, as shown in Fig.1, the MBS is connected to the nearby edge computing platform by optical fiber, and thus the MBS can easily get sufficient computing power support.

### 3.2 mmWave/THz Signal Propagation Model

In this subsection, we introduce the relevant radio signal propagation model in mmWave and THz frequency bands, and adopt the beam pattern proposed in [28,29] to measure beamforming gains, which is given by

$$G(\varphi, \vartheta) = \begin{cases} \frac{2\pi - (2\pi - \varphi)\epsilon}{\varphi}, & if \ |\vartheta| \leq \frac{\varphi}{2}, \\ \epsilon, & otherwise, \end{cases} \quad (1)$$

where $\varphi$ is the main lobe beam width in radian, and $\vartheta$ is the beam offset angle to the main lobe in radian; $\epsilon$ denotes a side lobe gain, which is a small positive number. Because of the high line-of-sight (LoS) probability of short mmWave/THz links, the directional transmission gain $G_{i,j}^{Tx}$ and the directional reception gain $G_{i,j}^{Rx}$ from node $i$ to node $j$ in the case of beam alignment can be derived by

$$\begin{cases} G_{i,j}^{Tx} = \frac{2\pi - (2\pi - \varphi_{i,j}^{Tx})\epsilon}{\varphi_{i,j}^{Tx}}, \\ G_{i,j}^{Rx} = \frac{2\pi - (2\pi - \varphi_{i,j}^{Rx})\epsilon}{\varphi_{i,j}^{Rx}}, \end{cases} \quad (2)$$

Where $\varphi_{i,j}^{Tx}$ is the transmitting beam width and $\varphi_{i,j}^{Rx}$ is the receiving beam width. Based on (2) and the mmWave/THz channel propagation model described in [30], the received power spectral density (PSD) in mmWave/THz links can be estimated by

$$PSD_{i,j}^{Rx}(f, d_{i,j}) = \frac{PSD_{i,j}^{Tx}(f) \, G_{i,j}^{Tx} \, G_{i,j}^{Rx}}{L_a(f, d_{i,j}) L_p(f, d_{i,j})} \quad (3)$$

where $PSD_{i,j}^{Rx}(f, d_{i,j})$ and $PSD_{i,j}^{Tx}(f)$ denote the received and the transmitted PSD values respectively, while $L_a(f, d_{i,j})$ and

$L_p(f, d_{i,j})$ denote the absorption loss and propagation loss respectively; $f$ denotes the operating frequency. According to the reference [31], the absorption loss can be evaluated by

$$L_a(f, d_{i,j}) \approx \frac{1}{e^{-K(f) d_{i,j}}} \quad (4)$$

where $K(f)$ is the overall absorption coefficient of the medium available from the HITRAN database [32]. In addition, under the assumption of spherical propagation in free space, the propagation loss can be evaluated by

$$L_p(f, d_{i,j}) = \left(\frac{4\pi d_{i,j}}{c}\right)^2 \quad (5)$$

where $c$ denotes the speed of light. Given mmWave/THz channel bandwidth $\mathcal{b}$, the transmitted power $p_{i,j}^t$ and received power $p_{i,j}^r$ can be evaluated by

$$\begin{cases} p_{i,j}^t = \mathcal{b} \cdot PSD_{i,j}^{Tx}(f, d_{i,j}), & (6a) \\ p_{i,j}^r = \mathcal{b} \cdot PSD_{i,j}^{Rx}(f, d_{i,j}), & (6b) \end{cases} \quad (6)$$

We take the two concurrent mmWave/THz links (e.g., $i \to j$ and $\hat{\imath} \to \hat{\jmath}$) for an example to derive the interference PSD and power perceived at $j$, which is expressed by

$$\begin{cases} PSD_{\hat{\imath}, \hat{\jmath} \to i,j}^{Rx} = \frac{PSD_{\hat{\imath}, \hat{\jmath} \to i,j}^{Tx}(f) G_{\hat{\imath}, \hat{\jmath} \to i,j}^{Tx} G_{\hat{\imath}, \hat{\jmath} \to i,j}^{Rx}}{L_a(f, d_{\hat{\imath}, j}) L_p(f, d_{\hat{\imath}, j})}, & (7a) \\ p_{\hat{\imath}, \hat{\jmath} \to i,j}^r = \mathcal{b} \cdot PSD_{\hat{\imath}, \hat{\jmath} \to i,j}^{Rx}, & (7b) \end{cases} \quad (7)$$

where $G_{\hat{\imath}, \hat{\jmath} \to i,j}^{Tx}$ and $G_{\hat{\imath}, \hat{\jmath} \to i,j}^{Rx}$ represent the directional transmission gain and directional reception gain of the link $\hat{\imath} \to \hat{\jmath}$, respectively. In addition, let $\vartheta_{\hat{\imath}, \hat{\jmath} \to i,j}^{Tx}$ be the offset angle between node $\hat{\imath}$'s transmitting beam direction and the link from node $\hat{\imath}$ to node $j$, and also let $\vartheta_{\hat{\imath}, \hat{\jmath} \to i,j}^{Rx}$ be the offset angle between node $j$'s receiving beam direction and the link to node $j$ from node $\hat{\imath}$. When $|\vartheta_{\hat{\imath}, \hat{\jmath} \to i,j}^{Tx}| \leq \frac{\varphi_{\hat{\imath}, \hat{\jmath}}^{Tx}}{2}$ and $|\vartheta_{\hat{\imath}, \hat{\jmath} \to i,j}^{Rx}| \leq \frac{\varphi_{i,j}^{Rx}}{2}$, the beam of the interfering node (i.e., $\hat{\imath}$) is aligned with that of the interfered node (i.e., $j$). According to the formula (2), the directional transmission-reception gain of interference path in the case of beam alignment can be given by

$$G_{\hat{\imath}, \hat{\jmath} \to i,j}^{Tx} G_{\hat{\imath}, \hat{\jmath} \to i,j}^{Rx} = \frac{2\pi - (2\pi - \varphi_{\hat{\imath}, \hat{\jmath}}^{Tx})\epsilon}{\varphi_{\hat{\imath}, \hat{\jmath}}^{Tx}} \cdot \frac{2\pi - (2\pi - \varphi_{i,j}^{Rx})\epsilon}{\varphi_{i,j}^{Rx}} \quad (8)$$

### 3.3 Power Discretization and Frame Structure Design

Power control is an effective and simple important means to improve energy efficiency in wireless networks. In this paper, we consider discrete power control for all the BSs and UEs. Let $p_{max}^{bs}$ and $p_{max}^{ue}$ denote the maximum transmit power of each BS and each UE, respectively. Then, the BS power set after discretization is expressed by

$$\mathcal{P}^{bs} = \{0, \frac{1}{l^{bs}} p_{max}^{bs}, \dots, \frac{l^{bs}}{l^{bs}} p_{max}^{bs}, \dots, p_{max}^{bs}\} \quad (9)$$

where BS power levels are $L^{bs} + 1$ and $1 \leq l^{bs} \leq L^{bs}$. Similarly, the UE power set after discretization is expressed by

$$\mathcal{P}^{ue} = \{0, \frac{1}{l^{ue}} p_{max}^{ue}, \dots, \frac{l^{ue}}{l^{ue}} p_{max}^{ue}, \dots, p_{max}^{ue}\} \quad (10)$$

where UE power levels are $L^{ue} + 1$ and $1 \leq l^{ue} \leq L^{ue}$. Since we mainly focus on using the mmWave band to transmit large capacity and long data streams, the number of UEs associated with each SBS at most $K$, where $\mathcal{U}^m = \{u_1^m, \dots, u_k^m, \dots, u_K^m\}$, $\forall m_m \in \mathcal{M}^-$. The set of mmWave spectrum resource blocks is given by $\mathcal{B}^{mm} = \{\mathcal{b}_1^{mm}, \dots, \mathcal{b}_n^{mm}, \dots, \mathcal{b}_N^{mm}\}$.

The seven UL/DL subframe configuration patterns specified by TD-LTE are used in this paper (see Fig.2), which is denoted by the set $\mathcal{D}$. Let $d_m \in \{0, 1, \ldots, 6\}$ represent the RFC selected by SBS $m_m$. Let $D = \{d_m\}_{\forall m_m \in \mathcal{M}^-}$ represent the set of RFCs selected by all the SBSs. As aforesaid, each frame length is 10ms and it includes 10 successive subframes with 1ms. In addition, each subframe with type S consists of downlink pilot timeslot, guard period, and the uplink pilot timeslot. Since the downlink pilot timeslot is allocated almost all of the orthogonal frequency division multiplexing symbols, it is assumed that a subframe with type S can be regarded as a subframe with type D for simplicity. For each subframe with type U or D, it is divided into radio access (RA) transmission duration and the another for radio backhaul (BH) transmission duration based on IAB architecture [4].

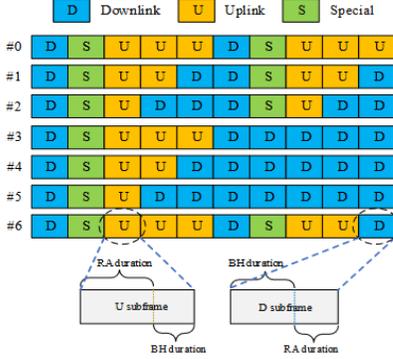

Fig. 2. TDD Frame structure design for UL/DL with RA/BH

### 3.4 Estimation Models for Interference and Throughput

When the D-TDD scheme [33] is used for each SBS to adapt to the asymmetric UL/DL traffic demand and the non-unified transmission duration allocation scheme [34] is adopted to balance the RA/BH traffic of each SBS, two new types of interferences (i.e., UE-UE CLI and SBS-SBS CLI) need to be considered besides the types of interferences concerned in [34].

In this paper, we adopt a clustering method to put the UEs with similar UL/DL traffic ratio into one cluster, which facilitates the association with the SBS that uses the same or similar ratio configuration in terms of subframe types. The characteristics of mmWave and THz communications, such as high path-loss, narrow beam width and sensitivity to blockage, will be beneficial in mitigating all the types of interferences, and thus make the non-unified transmission duration allocation scheme feasible.

Since each subframe consists of multiple time slots, we assume that each subframe can be divided into $Z$ time slots. For any SBS $m_m \in \mathcal{M}^-$ at the $t$-th frame, the RA/BH transmission duration during the $\tau$-th subframe after normalization satisfies the following relationship.

$$\gamma_{t,\tau}^m \in \Omega = \left\{ \frac{1}{Z}, \frac{2}{Z}, \ldots, \frac{Z-1}{Z} \right\} \quad (11)$$

Let $\Omega_{t,\tau} = \{\gamma_{t,\tau}^1, \ldots, \gamma_{t,\tau}^m, \ldots, \gamma_{t,\tau}^M\}$ be the RA/BH transmission duration sequence during the $\tau$-th subframe of the $t$-th frame corresponding to $M$ SBSs. For the convenience of analysis, the elements in $\Omega_{t,\tau}$ are rearranged in ascending order, and thus relabeled as $\widehat{\Omega}_{t,\tau} = \{\hat{\gamma}_{t,\tau}^1, \ldots, \hat{\gamma}_{t,\tau}^m, \ldots, \hat{\gamma}_{t,\tau}^M\}$. Moreover, let $L_{t,\tau} = \{L_{t,\tau}^1, \ldots, L_{t,\tau}^m, \ldots, L_{t,\tau}^M\}$ be the index sequence corresponding to an ascending sequence $\widehat{\Omega}_{t,\tau}$, where $L_{t,\tau}^m$ is the index of the $m$-th SBS. For the convenience of the following analysis, we specify that $\hat{\gamma}_{t,\tau}^0 = 0$ and $\hat{\gamma}_{t,\tau}^{M+1} = 1$.

In the resource allocation system joint considering dynamic UL/DL traffic ratio and non-unified RA/BH transmission duration, the four kinds of interfering sources may be considered to calculate the signal to interference and noise ratio (SINR) of each link. These interfering sources appear in RA and BH transmission durations of each U subframe, as well as BH and RA transmission durations of each D subframe, respectively. Since this paper mainly uses mmWave links for data transmission, supplemented by THz links, the following formulas for estimating interference and throughput are based on mmWave links by default, unless otherwise stated.

● Estimation of interference, SINR, and throughput in Case 1 in Fig.1

In Case 1 in Fig.1, for an UL access link from UE $u_k^m \in \mathcal{U}^m$ to SBS $m_m \in \mathcal{M}^-$ in the duration of $\hat{\gamma}_{t,\tau}^i - \hat{\gamma}_{t,\tau}^{i-1}$ for $1 \leq i \leq L_{t,\tau}^m$, the interference from which it will suffer may include 1) the interference from the UL and DL access links of all the SBSs except for those with indices in list $L_{t,\tau}$ being smaller than $i$, and 2) the interference from the UL and DL access links of all the SBSs with indices in $L_{t,\tau}$ being smaller than $i$. The above interference types are denoted by $I_{k,m}^{ra}$ and $I_{k,m}^{bh}$ respectively, which can be estimated by

$$I_{k,m}^{ra} = \sum_{\substack{m \in \\ \mathcal{M}^- \backslash \mathcal{M}_i^-}} \sum_{\substack{k \in \\ \mathcal{U}^{m} \backslash k}} \left( \begin{array}{c} (1 - D_{m}) \cdot PSD_{k, m \to k, m}^{Rx} \\ + D_{m} \cdot PSD_{k \to k, m}^{Rx} \end{array} \right) \cdot \mathcal{B} \quad (12)$$

and

$$I_{k,m}^{bh} = \sum_{m \in \mathcal{M}_i^-} \left( \begin{array}{c} (1 - D_{m}) \cdot PSD_{m, m_0 \to k, m}^{Rx} \\ + D_{m} \cdot PSD_{m_0, m \to k, m}^{Rx} \end{array} \right) \cdot \mathcal{B} \quad (13)$$

where $\mathcal{M}_i^-$ denotes the set of the SBSs with indices in $L_{t,\tau}$ being smaller than $i$ and $D_{m}$ is a binary indicator variable. If SBS $m_{m}$ is processing U subframe, $D_{m} = 0$. Otherwise, $D_{m} = 1$. Similarly, we can derive the interference estimation formulas of the corresponding DL access link from SBS $m_m$ to UE $u_k^m$, which is given by

$$I_{m,k}^{ra} = \sum_{\substack{m \in \\ \mathcal{M}^- \backslash \mathcal{M}_i^-}} \sum_{\substack{k \in \\ \mathcal{U}^{m} \backslash k}} \left( \begin{array}{c} (1 - D_{m}) \cdot PSD_{m, m \to m, k}^{Rx} \\ + D_{m} \cdot PSD_{m, k \to m, k}^{Rx} \end{array} \right) \cdot \mathcal{B} \quad (14)$$

and

$$I_{m,k}^{bh} = \sum_{m \in \mathcal{M}_i^-} \left( \begin{array}{c} (1 - D_{m}) \cdot PSD_{m, m_0 \to m, k}^{Rx} \\ + D_{m} \cdot PSD_{m_0, m \to m, k}^{Rx} \end{array} \right) \cdot \mathcal{B} \quad (15)$$

In Case 1 in Fig.1, for an UL backhaul link from SBS $m_m$ to MBS $m_0$ in the duration of $\hat{\gamma}_{t,\tau}^{i+1} - \hat{\gamma}_{t,\tau}^i$ for $L_{t,\tau}^m \leq i \leq M$, the interference from which it will suffer may include 1) the interference from the UL and DL access links of all the SBSs except for those with indices in list $L_{t,\tau}$ being smaller than $i$, and 2) the interference from the UL and DL backhaul links of all the SBSs with indices in $L_{t,\tau}$ being more than $i$. The above interference types are denoted by $I_{m,m_0}^{ra}$ and $I_{m,m_0}^{bh}$ respectively, which can be estimated by

$$I_{m,m_0}^{ra} = \sum_{\substack{m \in \\ \mathcal{M}^- \backslash \mathcal{M}_i^-}} \sum_{\substack{k \in \\ \mathcal{U}^{m} \backslash k}} \left( \begin{array}{c} (1 - D_{m}) \cdot PSD_{m, m \to m, m_0}^{Rx} \\ + D_{m} \cdot PSD_{m, k \to m, m_0}^{Rx} \end{array} \right) \cdot \mathcal{B} \quad (16)$$

and

$$I_{m,m_0}^{bh} = \sum_{m \in \mathcal{M}_i^- \backslash m} \left( \begin{array}{c} (1 - D_{m}) \cdot PSD_{m, m_0 \to m, m_0}^{Rx} \\ + D_{m} \cdot PSD_{m_0, m \to m, m_0}^{Rx} \end{array} \right) \cdot \mathcal{B} \quad (17)$$

Similarly, we can derive the interference estimation formulas of the corresponding DL backhaul link from MBS $m_0$ to SBS $m_m$, which is given by

$$I_{m_0,m}^{ra} = \sum_{\substack{m` \in \\ \mathcal{M}` \setminus \mathcal{M}_i^-}} \sum_{\substack{k` \in \\ \mathcal{U}^{m`} \setminus k}} \begin{pmatrix} (1 - D_{m`}) \cdot PSD_{k`,m` \to \atop m_0,m}^{Rx} \\ + D_{m`} \cdot PSD_{m`,k` \to m_0,m}^{Rx} \end{pmatrix} \cdot \mathcal{B} \quad (18)$$

and

$$I_{m_0,m}^{bh} = \sum_{m` \in \mathcal{M}_i^- \setminus m} \begin{pmatrix} (1 - D_{m`}) \cdot PSD_{m`,m_0 \to m_0,m}^{Rx} \\ + D_{m`} \cdot PSD_{m_0,m` \to m_0,m}^{Rx} \end{pmatrix} \cdot \mathcal{B} \quad (19)$$

The SINR estimation formulas for the UL and DL access links between UE $u_k^m$ and SBS $m_m$ in the duration of $\hat{\gamma}_{t,\tau}^i - \hat{\gamma}_{t,\tau}^{i-1}$ for $1 \leq i \leq L_{t,\tau}^m$ are estimated by

$$\begin{cases} SINR_{k,m}^i = \dfrac{\mathcal{B} \cdot PSD_{k,m}^{Rx}}{I_{k,m}^{ra} + I_{k,m}^{bh} + \mathcal{B} N_0}, & (20a) \\[2mm] SINR_{m,k}^i = \dfrac{\mathcal{B} \cdot PSD_{m,k}^{Rx}}{I_{m,k}^{ra} + I_{m,k}^{bh} + \mathcal{B} N_0}, & (20b) \end{cases} \quad (20)$$

where $N_0$ represents the background noise power spectrum density. Similarly, the SINR estimation formulas for the UL and DL backhaul links between MBS $m_0$ and SBS $m_m$ in the duration of $\hat{\gamma}_{t,\tau}^{i+1} - \hat{\gamma}_{t,\tau}^i$ for $L_m^m \leq i \leq M$ are estimated by

$$\begin{cases} SINR_{m,m_0}^i = \dfrac{\mathcal{B} \cdot PSD_{m,m_0}^{Rx}}{I_{m,m_0}^{ra} + I_{m,m_0}^{bh} + \mathcal{B} N_0}, & (21a) \\[2mm] SINR_{m_0,m}^i = \dfrac{\mathcal{B} \cdot PSD_{m_0,m}^{Rx}}{I_{m_0,m}^{ra} + I_{m_0,m}^{bh} + \mathcal{B} N_0}, & (21b) \end{cases} \quad (21)$$

The throughput estimation formulas for the UL and DL access links between UE $u_k^m$ and SBS $m_m$ can be given by

$$\begin{cases} T_{k,m}^{ra} = \mathcal{B} \sum_{i=1}^{L_{t,\tau}^m} \begin{pmatrix} (\hat{\gamma}_{t,\tau}^i - \hat{\gamma}_{t,\tau}^{i-1}) \cdot \\ \log_2(1 + SINR_{k,m}^i) \end{pmatrix}, & (22a) \\[3mm] T_{m,k}^{ra} = \mathcal{B} \sum_{i=1}^{L_{t,\tau}^m} \begin{pmatrix} (\hat{\gamma}_{t,\tau}^i - \hat{\gamma}_{t,\tau}^{i-1}) \cdot \\ \log_2(1 + SINR_{m,k}^i) \end{pmatrix}, & (22b) \end{cases} \quad (22)$$

Similarly, the throughput estimation formulas for the UL and DL backhaul links between MBS $m_0$ and SBS $m_m$ can be given by

$$\begin{cases} T_{m,m_0}^{bh} = \mathcal{B} \sum_{i=L_{t,\tau}^m}^{M} \begin{pmatrix} (\hat{\gamma}_{t,\tau}^{i+1} - \hat{\gamma}_{t,\tau}^i) \cdot \\ \log_2(1 + SINR_{m,m_0}^i) \end{pmatrix}, & (23a) \\[3mm] T_{m_0,m}^{bh} = \mathcal{B} \sum_{i=L_{t,\tau}^m}^{M} \begin{pmatrix} (\hat{\gamma}_{t,\tau}^{i+1} - \hat{\gamma}_{t,\tau}^i) \cdot \\ \log_2(1 + SINR_{m_0,m}^i) \end{pmatrix}, & (23b) \end{cases} \quad (23)$$

● Estimation of interference, SINR, and throughput in Case 2 in Fig.1

In Case 2 in Fig.1, there are two tandem backhaul links, where the relationship between the access links and the adjacent backhaul link is similar to Case 1. Therefore, the formulas in Case 1 can be used to estimate the interference and throughput in the corresponding links in Case 2. However, the relaying SBS must process the data transmission of its own associated UEs during the backhaul time so as not to affect the execution of the relay task. For the same reason, the backhaul link that is not adjacent to the access links must be activated during the access time for data transmission. The above time limit is based on the assumption that the same transmission path (composed of multiple links in series) multiplexes the same spectrum resource block. If additional spectrum resource blocks are allowed, the use of any exclusive spectrum resource block is not subject to the above time limit, but both ends of a communication link are required to have enough RF chains matching with additional spectrum resource blocks.

In order to save spectrum resources and improve spectrum utilization, additional spectrum resource blocks on the same transmission path will also be reused by other transmission paths. How do we properly divide the use time of additional spectrum resources to completely avoid interference? This will further complicate the problem modeling process, which even makes the problem unsolvable. Therefore, in order to simplify the problem and make use of the estimation formulas in Case 1 to estimate the interference and throughput achieved by the links using the additional spectrum resource blocks to transmit data, we stipulate that each relay SBS must strictly follow the access (or backhaul) transmission duration divided by the non-relay SBS on the same transmission path to transmit data.

● Estimation of interference, SINR, and throughput in Case 3 in Fig.1

Case 3 has one more relay SBS and one more backhaul link per transmission path than Case 2. Based on the reasons explained earlier, the newly added relay SBS must process its associated UEs' data during the access time. Also, as mentioned earlier, the access time is usually less than the backhaul time under the fixed scheduling period, so the actual data transmission time of the UEs associated with this type of relay SBS is shorter than that of the UEs associated with the other type of relay SBSs. The appropriate compensation of the loss in transmission time can be made by taking advantage of the high rate of THz band. As explained earlier, we should also increase the utilization of THR spectrum resource block by allowing it to be reused by the relay SBSs on the other transmission paths under Case 3. The formulas derived in Case 1 can be used to estimate the interference and throughput of mmWave links in Case 3. Therefore, we just need to give the estimation formulas of the interference and throughput of THz links as follows.

In Case 3 in Fig.1, for an UL access THz link from UE $u_k^m$ to SBS $m_m$ in the duration of $\hat{\gamma}_{t,\tau}^i - \hat{\gamma}_{t,\tau}^{i-1}$ for $1 \leq i \leq L_{t,\tau}^m$, it will only receive the interference from the UL and DL access THz links of all the relay SBSs except for those with indices in list $L_{t,\tau}$ being smaller than $i$.

$$I_{k,m}^{thz} = \sum_{\substack{m` \in \\ \mathcal{M}` \setminus \mathcal{M}_i^-}} \sum_{\substack{k` \in \\ \mathcal{U}^{m`} \setminus k}} \begin{pmatrix} (1 - D_{m`}) \cdot PSD_{k`,m` \atop \to k,m}^{Rx} \\ + D_{m`} \cdot PSD_{m`,k` \to k,m}^{Rx} \end{pmatrix} \cdot \mathcal{B}^{thz} \quad (24)$$

where $\mathcal{B}^{thz}$ denotes the bandwidth of one THz spectrum resource block. Similarly, we can derive the interference estimation formulas of the corresponding DL access THz link from SBS $m_m$ to UE $u_k^m$, which is given by

$$I_{m,k}^{thz} = \sum_{\substack{m` \in \\ \mathcal{M}` \setminus \mathcal{M}_i^-}} \sum_{\substack{k` \in \\ \mathcal{U}^{m`} \setminus k}} \begin{pmatrix} (1 - D_{m`}) \cdot PSD_{k`,m` \atop \to m,k}^{Rx} \\ + D_{m`} \cdot PSD_{m`,k` \to m,k}^{Rx} \end{pmatrix} \cdot \mathcal{B}^{thz} \quad (25)$$

The SINR estimation formulas for the UL and DL access THz links between UE $u_k^m$ and SBS $m_m$ in the duration of $\hat{\gamma}_{t,\tau}^i - \hat{\gamma}_{t,\tau}^{i-1}$ for $1 \leq i \leq L_{t,\tau}^m$ are estimated by

$$\begin{cases} SINR_{k,m}^{i,thz} = \dfrac{\mathcal{B}^{thz} \cdot PSD_{k,m}^{Rx}}{I_{k,m}^{thz} + \mathcal{B}_{thz} N_0}, & (26a) \\[2mm] SINR_{m,k}^{i,thz} = \dfrac{\mathcal{B}^{thz} \cdot PSD_{m,k}^{Rx}}{I_{m,k}^{thz} + \mathcal{B}_{thz} N_0}, & (26b) \end{cases} \quad (26)$$

The throughput estimation formulas for the UL and DL access THz links between UE $u_k^m$ and SBS $m_m$ can be given by

$$
\begin{cases}
T_{k,m}^{thz} = \mathcal{E}^{\cdot thz} \sum_{i=1}^{i_{t,\tau}^m} \begin{pmatrix} (\hat{y}_{t,\tau}^i - \hat{y}_{t,\tau}^{i-1}) \cdot \\ \log_2(1 + SINR_{k,m}^{i,thz}) \end{pmatrix}, & (27a) \\
\\
T_{m,k}^{thz} = \mathcal{E}^{\cdot thz} \sum_{i=1}^{i_{t,\tau}^m} \begin{pmatrix} (\hat{y}_{t,\tau}^i - \hat{y}_{t,\tau}^{i-1}) \cdot \\ \log_2(1 + SINR_{m,k}^{i,thz}) \end{pmatrix}, & (27b)
\end{cases} \quad (27)
$$

## 4. PROBLEM STATEMENT

In this paper, we want to jointly consider relay SBS selection, user association, discrete power control, and non-unified RA/BH transmission duration allocation of different UL/DL traffic ratios to maximize the throughput of the whole network on the premise of meeting the QoS requirements of all the concurrent UEs. It is very challenging to solve this problem due to the interweaving of multi-dimensional factors and multiple constraints. In order to reduce the difficulty of solving the problem, we give the following step-by-step solution ideas based on problem decomposition.

### 4.1 Relay SBS Selection and User Association

● Relay SBS selection problem.

Firstly, we need to generate a spanning tree rooted in MBS $m_0$ based on the IAB network consisting of all the BSs, and then regard the non-leaf nodes and leaf nodes in this spanning tree as the relay SBSs and non-relay SBSs, respectively. If each BS periodically reports its local link status (e.g., mmWave link quality with adjacent BS) to MBS $m_0$ via sub-6 GHz frequency band, MBS $m_0$ can always get a view of the IAB network in time and thus update the spanning tree by using Kruskal algorithm or one of classical spanning tree algorithms. At this point, the relay SBS selection problem is solved.

● User association problem.

From the updated spanning tree, we can clearly see each data transmission path between MBS $m_0$ and each non-relay SBS and thus know the number of relay SBSs on each transmission path. The more relay SBSs on a transmission path, the more difficult it is to guarantee end-to-end QoS. Therefore, in this paper, we will ensure that the number of SBSs on each transmission path does not exceed 3 by pruning the above spanning tree, which is based on the tradeoff between the coverage of mmWave IAB networks and the cost of ensuring end-to-end QoS.

In this paper, we mainly consider end-to-end throughput assurance between each UE and MBS $m_0$. Due to the aggregation characteristic of spanning tree, the closer the node is to the root, the more traffic is aggregated. In addition, by reducing the upper bound of throughput commitment of a SBS to UEs, the UEs with high throughput requirements can be excluded and thus the total throughput can be reduced under the same number of concurrent requests of this SBS. Therefore, for a transmission path, each SBS's throughput commitment upper bound should be based on its hop count from it to MBS $m_0$, which is expressed by

$$
T_m^{up} = \frac{T_{max}^{thd}}{H_m}, \ m_m \in \mathcal{M}^- \quad (28)
$$

where $T_{max}^{thd}$ represents the throughput commitment meeting the throughput requirement of all the UEs, $T_m^{up}$ represents the upper bound of throughput commitment of SBS $m_m$, and $H_m$ is the hop count from SBS $m_m$ to MBS $m_0$.

For SBS $m_m$ at the $t$-th time frame, where a time frame is also called a scheduling period, the weight of UE $u_k^m \in \mathcal{U}^m$ is defined by

$$
w_k^m(t) = \frac{1}{\hat{T}_k^m(t)} \quad (29)
$$

where $\hat{T}_k^m(t)$ represents the long-term average throughput of UE $u_k^m$ at the $t$-th time frame, which is defined by

$$
\hat{T}_k^m(t) = (1 - \delta_T)\hat{T}_k^m(t-1) + \delta_T T_k^m(t-1) \quad (30)
$$

where $\delta_T \in (0,1)$ represents a parameter close to zero, which specifies the window size for the exponential moving average operation. $T_k^m(t-1)$ is the throughput of UE $u_k^m$ at the $(t-1)$-th time frame, which is estimated by

$$
T_k^m(t-1) = \sum_{\tau=1}^{\mathcal{L}} \frac{D_m^\tau(t-1)T_{k,m}^{\tau a} + (1 - D_m^\tau(t-1))T_{m,k}^{\tau a}}{\mathcal{L}} \quad (31)
$$

where $D_m^\tau(t-1)$ is equal to 0 if SBS $m_m$ is processing UL traffic during $\tau$-th subframe of the $(t-1)$-th time frame, otherwise it is equal to 1. When the number (denoted by $K_{us}^m$, $K_{us}^m \le K$) of concurrent services of SBS $m_m$ is less than the cardinality $|\mathcal{U}^m|$ of the set $\mathcal{U}^m$, the top-$K_{us}^m$ UEs need to be selected from the set $\mathcal{U}^m$. The proportional-fairness (PF) ratio defined in [35] is adopted to select the top-$K_{us}^m$ UEs, which is given by

$$
PF_k^m(t) = w_k^m(t) \log_2(1 + \widetilde{SINR}_k^m(t)) \quad (32)
$$

where $\widetilde{SINR}_k^m(t)$ represents the long-term average SINR of the link between UE $u_k^m$ and SBS $m_m$ at the $t$-th time frame, which is defined by

$$
\widetilde{SINR}_k^m(t) = (1 - \delta_S)\widetilde{SINR}_k^m(t) + \delta_S SINR_k^m(t-1) \quad (33)
$$

where $\delta_S \in (0,1)$ represents a parameter close to zero, which specifies the window size for the exponential moving average operation. $SINR_k^m(t-1)$ is the SINR of the link between UE $u_k^m$ and SBS $m_m$ at the $t$-th time frame, which is estimated by

$$
SINR_k^m(t-1) = \sum_{\tau=1}^{\mathcal{L}} \frac{SINR_{m,k}^L + SINR_{k,m}^L}{\mathcal{L}} \quad (34)
$$

where $SINR_{m,k}^L$ and $SINR_{k,m}^L$ are detailed as follows.

$$
\begin{cases}
SINR_{m,k}^L = \sum_{i=1}^{i_{t,\tau}^m} \begin{pmatrix} D_m^\tau(t-1) \cdot \\ (\hat{y}_{t,\tau}^i - \hat{y}_{t,\tau}^{i-1}) \cdot SINR_{m,k}^i \end{pmatrix} & (35a) \\
\\
SINR_{k,m}^L = \sum_{i=1}^{i_{t,\tau}^m} \begin{pmatrix} (1 - D_m^\tau(t-1)) \cdot \\ (\hat{y}_{t,\tau}^i - \hat{y}_{t,\tau}^{i-1}) \cdot SINR_{k,m}^i \end{pmatrix} & (35b)
\end{cases} \quad (35)
$$

To reduce the complexity of solving the problem, we stipulate that a non-relay SBS serves only one type of UEs with a given UL/DL traffic ratio. In an ultra-dense network, UEs with different UL/DL traffic ratios can always find a base station to get access service. To prevent the overlap of partial links on transmission paths serving UEs with different UL/DL traffic ratios, the number of concurrent transmission paths is at most the number of RF chains at MBS $m_0$. In fact, a long transmission path may occupy multiple RF chains of MBS $m_0$ for concurrent transmission. This, in turn, makes it easier to find disjoint and non-overlapping concurrent transmission paths on the spanning tree.

How to allocate the RF chains of MBS $m_0$ is directly related to the determination of the number of concurrent transmission paths. Let the number of concurrent transmission paths be denoted by $K_p^0$. When $K_p^0$ is less than the number of non-relay SBSs for which UL/DL traffic ratios have been determined, the top-$K_p^0$ non-relay SBSs need to be selected. Similar to the formula (33), we define the following formula to proportional-fairly select the top-$K_p^0$ concurrent transmission paths.

$$PF_{pl}^0(t) = \frac{\sum_{pl \in PL} T_{pl}^0(t)}{\tilde{T}_{pl}^0(t)} \qquad (36)$$

where $PL$ is the set of concurrent non-relay SBSs and $|PL| = K_p^0$, and $\tilde{T}_{pl}^0(t)$ represents the long-term average throughput of non-relaying SBS of $pl$-th transmission path at the $t$-th time frame, which is defined by

$$\tilde{T}_{pl}^0(t) = (1 - \delta_p)\tilde{T}_{pl}^0(t-1) + \delta_P T_{pl}^0(t-1) \qquad (37)$$

where $\delta_p \in (0,1)$ represents a parameter close to zero, which specifies the window size for the exponential moving average operation. $T_{pl}^0(t-1)$ is the throughput of non-relay SBS of $pl$-th transmission path at the $(t-1)$-th time frame, which is estimated by

$$T_{pl}^0(t-1) = \sum_{k=1}^{K_{us}^{pl}} (T_{k,pl}^{ra} + T_{pl,k}^{ra}) \qquad (38)$$

where $K_{us}^{pl} = K_{us}^m$, $T_{k,pl}^{ra} = T_{k,m}^{ra}$, and $T_{pl,k}^{ra} = T_{m,k}^{ra}$ if $pl = m$. At this point, the user association problem is solved.

### 4.2 Discrete Power Control and Non-Unified RA/BH Transmission Duration Allocation Under Different UL/DL Traffic Ratios

Based on the above results of relay SBS selection and user association, we need to first derive the total throughput, total power consumption, and total spectrum consumption of the whole network, and then conveniently express the spectrum energy efficiency optimization problem.

We use $PL^1$ to denote the set of transmission paths in Case 1, $PL^2$ to denote the set of transmission paths in Case 2, and $PL^3$ to denote the set of transmission paths in Case 3. In addition, for the $j$-th transmission path $pl_j^1 \in PL^1$, $1 \leq j \leq |PL^1|$, $pl_j^1$ also represents the only SBS on the transmission path. However, for the $j$-th transmission path $pl_j^2 \in PL^2$, $1 \leq j \leq |PL^2|$, $pl_{j,1}^2$ denotes the non-relay SBS on the transmission path while $pl_{j,2}^2$ denotes the relay SBS on the transmission path. Similarly, for the $j$-th transmission path $pl_j^3 \in PL^3$, $1 \leq j \leq |PL^3|$, $pl_{j,1}^3$ denotes the non-relay SBS on the transmission path, $pl_{j,2}^3$ denotes the relay SBS adjacent to the non-relay SBS on the transmission path, and $pl_{j,3}^3$ denotes the relay SBS adjacent to MBS $m_0$. Based on the above symbolic definitions, we can easily express the UL/DL total throughput, total spectrum consumption, and total power consumption in three cases of the system as follows.

● **Uplink Scenario**

In Case 1, for a given transmission path set $PL^1$, the total UL access throughput $T_{PL^1}^{ura}$ is estimated by

$$T_{PL^1}^{ura} = \sum_{pl_j^1 \in PL^1} \sum_{k=1}^{K_{us}^{pl_j^1}} T_{k,pl_j^1}^{ra} \qquad (39)$$

where $T_{PL^1}^{ura}$ must satisfy the following constraint.

$$\sum_{k=1}^{K_{us}^{pl_j^1}} T_{k,pl_j^1}^{ra} \leq T_{pl_j^1,m_0}^{bh}, \forall pl_j^1 \in PL^1 \qquad (40)$$

where $T_{k,pl_j^1}^{ra}$ is computed by the formula (23a) via specifying the desired parameter values, while $T_{pl_j^1,m_0}^{bh}$ is estimated by the formula (24a) via specifying the desired parameter values.

Only one mmWave spectrum resource block is used in Case 1, but the total UL power consumption $P_{PL^1}^{ura}$ is estimated by

$$P_{PL^1}^{ura} = \sum_{pl_j^1 \in PL^1} \left( \sum_{k=1}^{K_{us}^{pl_j^1}} p_{k,pl_j^1}^{ue} + p_{pl_j^1,m_0}^{bs} \right) \qquad (41)$$

where $P_{PL^1}^{ura}$ must satisfy the following constraints.

$$\begin{cases} p_{k,pl_j^1}^{ue} \in \mathcal{P}^{ue} & (42a) \\ p_{pl_j^1,m_0}^{bs} \in \mathcal{P}_u^{sbs} \subset \mathcal{P}^{bs} & (42b) \end{cases} \qquad (42)$$

In Case 2, for a given transmission path set $PL^2$, the total UL access throughput $T_{PL^2}^{ura}$ is estimated by

$$T_{PL^2}^{ura} = \sum_{pl_j^2 \in PL^2} \left( \sum_{k=1}^{K_{us}^{pl_{j,1}^2}} T_{k,pl_{j,1}^2}^{ra} + \sum_{k=1}^{K_{us}^{pl_{j,2}^2}} T_{k,pl_{j,2}^2}^{ra} \right) \qquad (43)$$

where $T_{PL^2}^{ura}$ must satisfy the following constraints.

$$\begin{cases} \sum_{k=1}^{K_{us}^{pl_{j,1}^2}} T_{k,pl_{j,1}^2}^{ra} \leq T_{pl_{j,1}^2,pl_{j,2}^2}^{relay} \\ \left( \begin{matrix} \sum_{k=1}^{pl_{j,1}^2} T_{k,pl_{j,1}^2}^{ra} \\ + \sum_{k=1}^{pl_{j,2}^2} T_{k,pl_{j,2}^2}^{ra} \end{matrix} \right) \leq T_{pl_{j,2}^2,m_0}^{bh} \end{cases}, \forall pl_j^2 \in PL^2 \qquad (44)$$

where $T_{k,pl_{j,1}^2}^{ra}$ is computed by the formula (22a) via specifying the desired parameter values; $T_{pl_{j,1}^2,pl_{j,2}^2}^{relay}$ is estimated by the formula (23a) via specifying the desired parameter values; $T_{k,pl_{j,2}^2}^{ra}$ is computed by the formula (23a) via specifying the desired parameter values; $T_{pl_{j,2}^2,m_0}^{bh}$ is estimated by the formula (22a) via specifying the desired parameter values.

There may be several spectrum resource blocks used concurrently to achieve the throughput $T_{pl_{j,2}^2,m_0}^{bh}$, which is further expressed by

$$T_{pl_{j,2}^2,m_0}^{bh} = |\mathcal{B}_{PL^2}^{mm}| \mathcal{E}^{mm} \sum_{i=1}^{L_{t,\tau}^m} \left( \begin{matrix} (\hat{\gamma}_{t,\tau}^i - \hat{\gamma}_{t,\tau}^{i-1}) \cdot \\ \log_2\left( 1 + SINR_{pl_{j,2}^2,m_0}^i \right) \end{matrix} \right) \qquad (45)$$

where $\mathcal{B}_{PL^2}^{mm} \subset \mathcal{B}^{mm}$ is the set of spectrum resource blocks used in Case 2 and $\mathcal{E}^{mm}$ is the bandwidth of one mmWave spectrum resource block. Since more than one mmWave spectrum resource block may be used in Case 2, the total spectrum resource consumption is $|\mathcal{B}_{PL^2}^{mm}| \mathcal{E}^{mm}$. The total UL power consumption $P_{PL^2}^{ura}$ in Case 2 is estimated by

$$P_{PL^2}^{ura} = \sum_{pl_j^2 \in PL^2} \left( \begin{matrix} \sum_{k=1}^{K_{us}^{pl_{j,1}^2}} p_{k,pl_{j,1}^2}^{ue} + p_{pl_{j,1}^2,pl_{j,2}^2}^{bs} + \\ \sum_{k=1}^{K_{us}^{pl_{j,2}^2}} p_{k,pl_{j,2}^2}^{ue} + |\mathcal{B}_{PL^2}^{mm}| p_{pl_{j,2}^2,m_0}^{bs} \end{matrix} \right) \qquad (46)$$

where $P_{PL^2}^{ura}$ must satisfy the following constraints.

$$\begin{cases} p_{k,pl_{j,1}^2}^{ue}, p_{k,pl_{j,2}^2}^{ue} \in \mathcal{P}^{ue} & (47a) \\ p_{pl_{j,1}^2,pl_{j,2}^2}^{bs}, p_{pl_{j,2}^2,m_0}^{bs} \in \mathcal{P}_u^{sbs} & (47b) \end{cases} \qquad (47)$$

In Case 3, for a given transmission path set $PL^3$, the total UL access throughput $T_{PL^3}^{ura}$ is estimated by

$$T_{PL^3}^{ura} = \sum_{pl_j^3 \in PL^3} \begin{pmatrix} \sum_{k=1}^{K_{us}^{pl_{j,1}^3}} T_{k,pl_{j,1}^3}^{ra} + \sum_{k=1}^{K_{us}^{pl_{j,2}^3}} T_{k,pl_{j,2}^3}^{ra} \\ + \sum_{k=1}^{K_{us}^{pl_{j,3}^3}} T_{k,pl_{j,3}^3}^{ra} \end{pmatrix} \quad (48)$$

where $T_{PL^3}^{ura}$ must satisfy the following constraints.

$$\begin{cases} \sum_{k=1}^{K_{us}^{pl_{j,1}^3}} T_{k,pl_{j,1}^3}^{ra} \leq T_{pl_{j,1}^3,pl_{j,2}^3}^{relay} \\ \begin{pmatrix} \sum_{k=1}^{K_{us}^{pl_{j,1}^3}} T_{k,pl_{j,1}^3}^{ra} \\ + \sum_{k=1}^{K_{us}^{pl_{j,2}^3}} T_{k,pl_{j,2}^3}^{ra} \end{pmatrix} \leq T_{pl_{j,2}^3,pl_{j,3}^3}^{relay} \\ \begin{pmatrix} \sum_{k=1}^{K_{us}^{pl_{j,1}^3}} T_{k,pl_{j,1}^3}^{ra} \\ + \sum_{k=1}^{K_{us}^{pl_{j,2}^3}} T_{k,pl_{j,2}^3}^{ra} \\ + \sum_{k=1}^{K_{us}^{pl_{j,3}^3}} T_{k,pl_{j,3}^3}^{ra} \end{pmatrix} \leq T_{pl_{j,3}^3,m_0}^{bh} \end{cases}, \forall pl_j^3 \in PL^3 \quad (49)$$

where $T_{k,pl_{j,1}^3}^{ra}$ is computed by the formula (22a) via specifying the desired parameter values; $T_{pl_{j,1}^3,pl_{j,2}^3}^{relay}$ is estimated by the formula (23a) via specifying the desired parameter values; $T_{k,pl_{j,2}^3}^{ra}$ is computed by the formula (23a) via specifying the desired parameter values; $T_{pl_{j,2}^3,pl_{j,3}^3}^{relay}$ is estimated by the formula (22a) via specifying the desired parameter values; $T_{pl_{j,3}^3,m_0}^{bh}$ is estimated by the formula (23a) via specifying the desired parameter values; $T_{k,pl_{j,3}^3}^{ra}$ is computed by the formula (27a) via specifying the desired parameter values if one THz spectrum resource block is used, otherwise is computed by the formula (23a) via specifying the desired parameter values.

There may be several spectrum resource blocks used concurrently to achieve the throughput $T_{pl_{j,2}^3,pl_{j,3}^3}^{relay}$, which is further expressed by

$$T_{pl_{j,2}^3,pl_{j,3}^3}^{relay} = |\mathcal{B}_{PL^2}^{mm}| \mathcal{E}^{mm} \sum_{i=1}^{L_{t,\tau}^m} \begin{pmatrix} (\hat{\gamma}_{t,\tau}^i - \hat{\gamma}_{t,\tau}^{i-1}) \cdot \\ \log_2 \left(1 + SINR_{pl_{j,2}^3,pl_{j,3}^3}^i\right) \end{pmatrix} \quad (50)$$

Similarly, there may be several spectrum resource blocks used concurrently to achieve the throughput $T_{pl_{j,3}^3,m_0}^{bh}$, which is further expressed by

$$T_{pl_{j,3}^3,m_0}^{bh} = |\mathcal{B}_{PL^3}^{mm}| \mathcal{E}^{mm} \sum_{i=L_{t,\tau}^m}^{M} \begin{pmatrix} (\hat{\gamma}_{t,\tau}^{i+1} - \hat{\gamma}_{t,\tau}^i) \cdot \\ \log_2 \left(1 + SINR_{pl_{j,3}^3,m_0}^i\right) \end{pmatrix} \quad (51)$$

where $\mathcal{B}_{PL^3}^{mm}$ is the set of spectrum resource blocks used in Case 3 and $\mathcal{B}_{PL^2}^{mm} \subset \mathcal{B}_{PL^3}^{mm} \subset \mathcal{B}^{mm}$. The total mmWave spectrum resource consumption in Case 3 is $|\mathcal{B}_{PL^3}^{mm}| \mathcal{E}^{mm}$. If one THz spectrum resource block is used for achieving $T_{k,pl_{j,3}^3}^{ra}$, the total spectrum resource consumption is $|\mathcal{B}_{PL^3}^{mm}| \mathcal{E}^{mm} + \mathcal{E}^{thz}$. The total UL power consumption $P_{PL^3}^{ura}$ in Case 3 is estimated by

$$P_{PL^3}^{ura} = \sum_{pl_j^3 \in PL^3} \begin{pmatrix} \sum_{k=1}^{K_{us}^{pl_{j,1}^3}} p_{k,pl_{j,1}^3}^{ue} + p_{pl_{j,1}^3,pl_{j,2}^3}^{bs} + \\ \sum_{k=1}^{K_{us}^{pl_{j,2}^3}} p_{k,pl_{j,2}^3}^{ue} + |\mathcal{B}_{PL^2}^{mm}| \cdot p_{pl_{j,2}^3,pl_{j,3}^3}^{bs} + \\ \sum_{k=1}^{K_{us}^{pl_{j,3}^3}} p_{k,pl_{j,3}^3}^{ue} + |\mathcal{B}_{PL^3}^{mm}| \cdot p_{pl_{j,3}^3,m_0}^{bs} \end{pmatrix} \quad (52)$$

where $P_{PL^3}^{ura}$ must satisfy the following constraints.

$$\begin{cases} p_{k,pl_{j,1}^3}^{ue}, p_{k,pl_{j,2}^3}^{ue}, p_{k,pl_{j,3}^3}^{ue} \in \mathcal{P}^{ue} & (53a) \\ p_{pl_{j,1}^3,pl_{j,2}^3}^{bs}, p_{pl_{j,2}^3,pl_{j,3}^3}^{bs}, p_{pl_{j,3}^3,m_0}^{bs} \in \mathcal{P}_u^{sbs} & (53b) \end{cases} \quad (53)$$

● **Downlink Scenario**

Since UL and DL communications share the same spectrum resources in a TDD mode, we only need to express the DL total throughput and total power consumption in the three cases of the system.

In Case 1, for a given transmission path set $PL^1$, the total DL backhaul throughput $T_{PL^1}^{dbh}$ is estimated by

$$T_{PL^1}^{dbh} = \sum_{pl_j^1 \in PL^1} T_{m_0,pl_j^1}^{bh} \quad (54)$$

where $T_{PL^1}^{dbh}$ must satisfy the following constraint.

$$T_{m_0,pl_j^1}^{bh} \leq \sum_{k=1}^{K_{us}^{pl_j^1}} T_{pl_{j,1}^1,k}^{ra}, \forall pl_j^1 \in PL^1 \quad (55)$$

where $T_{pl_j^1,k}^{ra}$ is estimated by the formula (22b) via specifying the desired parameter values, while $T_{m_0,pl_j^1}^{bh}$ is computed by the formula (23b) via specifying the desired parameter values. The total DL power consumption $P_{PL^1}^{dbh}$ is estimated by

$$P_{PL^1}^{dbh} = \sum_{pl_j^1 \in PL^1} \begin{pmatrix} p_{m_0,pl_j^1}^{bs} + \sum_{k=1}^{K_{us}^{pl_j^1}} p_{pl_j^1,k}^{bs} \end{pmatrix} \quad (56)$$

where $P_{PL^1}^{dbh}$ must satisfy the following constraints.

$$\begin{cases} p_{m_0,pl_j^1}^{bs} \in \mathcal{P}^{mbs} \subset \mathcal{P}^{bs} & (57a) \\ p_{pl_j^1,k}^{bs} \in \mathcal{P}_d^{sbs} \subset \mathcal{P}^{bs} & (57b) \end{cases} \quad (57)$$

In Case 2, for a given transmission path set $PL^2$, the total DL backhaul throughput $T_{PL^2}^{dbh}$ is estimated by

$$T_{PL^2}^{dbh} = \sum_{pl_j^2 \in PL^2} T_{m_0,pl_{j,2}^2}^{bh} \quad (58)$$

where $T_{PL^2}^{dbh}$ must satisfy the following constraints.

$$\begin{cases} T_{m_0,pl_{j,2}^2}^{bh} \leq \begin{pmatrix} \sum_{k=1}^{K_{us}^{pl_{j,2}^2}} T_{pl_{j,1}^2,k}^{ra} \\ + T_{pl_{j,2}^2,pl_{j,1}^2}^{relay} \end{pmatrix} \\ T_{pl_{j,2}^2,pl_{j,1}^2}^{relay} \leq \sum_{k=1}^{K_{us}^{pl_{j,1}^2}} T_{pl_{j,1}^2,k}^{ra} \end{cases}, \forall pl_j^2 \in PL^2 \quad (59)$$

where $T_{pl_{j,2}^2,k}^{ra}$ is estimated by the formula (22b) via specifying the desired parameter values; $T_{pl_{j,2}^2,pl_{j,1}^2}^{relay}$ is estimated by the formula (23b) via specifying the desired parameter values; $T_{pl_{j,1}^2,k}^{ra}$ is estimated by the formula (23b) via specifying the desired parameter values; $T_{m_0,pl_{j,2}^2}^{bh}$ is computed by the formula (22b) via specifying the desired parameter values. When there may be several spectrum resource blocks used concurrently to achieve the throughput $T_{m_0,pl_{j,2}^2}^{bh}$, it is further expressed by

$$T^{bh}_{m_0,pl^2_{j,2}} = |\mathcal{B}^{mm}_{PL2}| \, \mathcal{E}^{mm} \sum_{i=1}^{L^m_{t,\tau}} \begin{pmatrix} (\hat{\gamma}^i_{t,\tau} - \hat{\gamma}^{i-1}_{t,\tau}) \cdot \\ \log_2\left(1 + SINR^i_{m_0,pl^2_{j,2}}\right) \end{pmatrix} \quad (60)$$

The total DL power consumption $P^{dbh}_{PL2}$ in Case 2 is estimated by

$$P^{dbh}_{PL2} = \sum_{pl^2_j \in PL2} \begin{pmatrix} \sum_{k=1}^{K^{pl^2_{j,1}}_{us}} p^{bs}_{pl^2_{j,1},k} + p^{bs}_{pl^2_{j,2},pl^2_{j,1}} + \\ \sum_{k=1}^{K^{pl^2_{j,2}}_{us}} p^{bs}_{pl^2_{j,2},k} + |\mathcal{B}^{mm}_{PL2}| \cdot p^{bs}_{m_0,pl^2_{j,2}} \end{pmatrix} \quad (61)$$

where $P^{dbh}_{PL2}$ must satisfy the following constraints.

$$\begin{cases} p^{bs}_{m_0,pl^2_{j,2}} \in \mathcal{P}^{mbs} & (62a) \\ p^{bs}_{pl^2_{j,2},pl^2_{j,1}}, p^{bs}_{pl^2_{j,2},k}, p^{bs}_{pl^2_{j,1},k} \in \mathcal{P}^{sbs}_d & (62b) \end{cases} \quad (62)$$

In Case 3, for a given transmission path set $PL^3$, the total DL backhaul throughput $T^{dbh}_{PL3}$ is estimated by

$$T^{dbh}_{PL3} = \sum_{pl^3_j \in PL3} T^{bh}_{m_0,pl^3_{j,3}} \quad (63)$$

where $T^{dbh}_{PL3}$ must satisfy the following constraints.

$$\begin{cases} T^{bh}_{m_0,pl^3_{j,3}} \leq \begin{pmatrix} \sum_{k=1}^{K^{pl^3_{j,3}}_{us}} T^{ra}_{pl^3_{j,3},k} \\ + T^{relay}_{pl^3_{j,3},pl^3_{j,2}} \end{pmatrix} \\ T^{relay}_{pl^3_{j,3},pl^3_{j,2}} \leq \begin{pmatrix} \sum_{k=1}^{pl^3_{j,2}} T^{ra}_{pl^3_{j,2},k} \\ + T^{relay}_{pl^3_{j,3},pl^3_{j,1}} \end{pmatrix}, \forall pl^3_j \in PL^3 \\ T^{relay}_{pl^3_{j,2},pl^3_{j,1}} \leq \sum_{k=1}^{pl^3_{j,1}} T^{ra}_{pl^3_{j,1},k} \end{pmatrix} \quad (64)$$

where $T^{ra}_{pl^3_{j,1},k}$ is estimated by the formula $(22b)$ via specifying the desired parameter values; $T^{relay}_{pl^3_{j,2},pl^3_{j,1}}$ is estimated by the formula $(23b)$ via specifying the desired parameter values; $T^{ra}_{pl^3_{j,2},k}$ is estimated by the formula $(23b)$ via specifying the desired parameter values; $T^{relay}_{pl^3_{j,3},pl^3_{j,2}}$ is estimated by the formula $(22b)$ via specifying the desired parameter values; $T^{bh}_{m_0,pl^3_{j,3}}$ is estimated by the formula $(23b)$ via specifying the desired parameter values; $T^{ra}_{pl^3_{j,3},k}$ is computed by the formula $(27b)$ via specifying the desired parameter values if one THz spectrum resource block is used, otherwise it is computed by the formula $(23b)$ via specifying the desired parameter values.

When there may be several spectrum resource blocks used concurrently to achieve the throughput $T^{relay}_{pl^3_{j,3},pl^3_{j,2}}$, it is further expressed by

$$T^{relay}_{pl^3_{j,3},pl^3_{j,2}} = |\mathcal{B}^{mm}_{PL2}| \, \mathcal{E}^{mm} \sum_{i=1}^{L^m_{t,\tau}} \begin{pmatrix} (\hat{\gamma}^i_{t,\tau} - \hat{\gamma}^{i-1}_{t,\tau}) \cdot \\ \log_2\left(1 + SINR^i_{pl^3_{j,3},pl^3_{j,2}}\right) \end{pmatrix} \quad (65)$$

Similarly, when there may be several spectrum resource blocks used concurrently to achieve the throughput $T^{bh}_{m_0,pl^3_{j,3}}$, it is further expressed by

$$T^{bh}_{m_0,pl^3_{j,3}} = |\mathcal{B}^{mm}_{PL3}| \, \mathcal{E}^{mm} \sum_{i=L^m_{t,\tau}}^{M} \begin{pmatrix} (\hat{\gamma}^{i+1}_{t,\tau} - \hat{\gamma}^i_{t,\tau}) \cdot \\ \log_2\left(1 + SINR^i_{m_0,pl^3_{j,3}}\right) \end{pmatrix} \quad (66)$$

The total DL power consumption $P^{dbh}_{PL3}$ in Case 3 is estimated by

$$P^{dbh}_{PL3} = \sum_{pl^3_j \in PL3} \begin{pmatrix} \sum_{k=1}^{pl^3_{j,1}} p^{bs}_{pl^3_{j,1},k} + p^{bs}_{pl^3_{j,2},pl^3_{j,1}} + \\ \sum_{k=1}^{pl^3_{j,2}} p^{bs}_{pl^3_{j,2},k} + |\mathcal{B}^{mm}_{PL2}| \cdot p^{bs}_{pl^3_{j,3},pl^3_{j,2}} + \\ \sum_{k=1}^{K^{pl^3_{j,3}}_{us}} p^{bs}_{pl^3_{j,3},k} + |\mathcal{B}^{mm}_{PL3}| \cdot p^{bs}_{m_0,pl^3_{j,3}} \end{pmatrix} \quad (67)$$

where $P^{dbh}_{PL3}$ must satisfy the following constraints.

$$\begin{cases} p^{bs}_{m_0,pl^3_{j,3}} \in \mathcal{P}^{mbs} & (68a) \\ p^{bs}_{pl^3_{j,3},pl^3_{j,2}}, p^{bs}_{pl^3_{j,2},pl^3_{j,1}}, p^{bs}_{pl^3_{j,3},k}, p^{bs}_{pl^3_{j,2},k}, p^{bs}_{pl^3_{j,1},k} \in \mathcal{P}^{sbs}_d & (68b) \end{cases} \quad (68)$$

### ● Problem Description for Joint UL and DL Scenarios

Based on the above results, the system throughput $T_q$ and the system power consumption $P_q$ can be expressed as follows.

$$\begin{cases} T_q = T^{ura}_{PL1} + T^{dbh}_{PL1} + T^{ura}_{PL2} + T^{dbh}_{PL2} + T^{ura}_{PL3} + T^{dbh}_{PL3} & (69a) \\ P_q = P^{ura}_{PL1} + P^{dbh}_{PL1} + P^{ura}_{PL2} + P^{dbh}_{PL2} + P^{ura}_{PL3} + P^{dbh}_{PL3} & (69b) \end{cases} \quad (69)$$

The system average spectral energy efficiency can be expressed as follows.

$$S_{EE} = \frac{T_q}{P_q \cdot \left(|\mathcal{B}^{mm}_{PL3}| \, \mathcal{E}^{mm} + \mathcal{E}^{thz}\right)} \quad (70)$$

The integrated optimization problem in terms of the system throughput is formulated by

$$\begin{cases} IOP: \max\limits_{\substack{PL^1,PL^2,PL^3,\mathcal{B}^{mm}_{PL2},\mathcal{B}^{mm}_{PL3} \\ \mathcal{P}^{ue},\mathcal{P}^{sbs}_u,\mathcal{P}^{sbs}_d,\mathcal{P}^{mbs},\mathcal{D},\Omega}} T_q \\ s.t. \quad (40),(42),(44),(47),(49),(53), \\ \qquad (55),(57),(59),(62),(64),(68) \\ C1: \mathcal{B}^{mm}_{PL2} \subset \mathcal{B}^{mm}_{PL3} \subset \mathcal{B}^{mm} \\ C2: T^{min}_{k,m} \leq T^{ra}_{k,m} \leq T^{up}_m, \forall m_m \in \mathcal{M}^-, \forall u^m_k \in \mathcal{U}^m \\ C3: T^{min}_{m,k} \leq T^{ra}_{m,k} \leq T^{up}_m, \forall m_m \in \mathcal{M}^-, \forall u^m_k \in \mathcal{U}^m \\ C4: S_{EE} \geq S^{cons}_{EE} \end{cases} \quad (71)$$

where the constraint $C1$ specifies that the available mmWave spectrum resource blocks; the constraints $C2$ and $C3$ specify that each UE should meet the UL and DL minimum throughput requirements (i.e., $T^{min}_{k,m}$ and $T^{min}_{m,k}$) respectively, and should not exceed the upper bound (i.e., $T^{up}_m$) on data rate that each SBS promises to provide; the constraint $C4$ specifies that the minimum system spectral efficiency which the system should meet.

## 5. PROBLEM SOLVING SCHEME

### 1) SL-HMF-RA Scheme

There are the nonlinear function form and multiple discrete variables in the IOP, so it is a mixed-integer nonlinear programming (MINLP) problem. Although it can be solved theoretically by an exhaustive method, its solution usually requires a huge search space with exponential time complexity [36]. Specifically, the IOP involves the six-dimensional parameters: the RFCs of SBSs (one-dimensional parameter), the RA/BH transmission duration division (one-dimensional parameter) and UL/DL discrete power allocation of SBSs (two-dimensional parameter), the discrete power allocation of UEs (one-dimensional parameter), and the DL discrete power allocation of the MBS to each SBS (one-dimensional parameter).

In order to reduce the time complexity of searching the suboptimal solution to the IOP, we decompose the IOP into the four sub-problems (i.e., $SP1$, $SP2$, $SP3$, $SP4$). The $SP1$ only focuses on the discrete power allocation optimization for UEs. The $SP2$ only concentrates on the optimization of RA/BH

| Notation | Meaning |
|---|---|
| $\mathcal{P}^{ue}$ | The set of transmission powers for UEs |
| $\mathcal{P}^{sbs}_u$ | The set of UL transmission powers for SBSs |
| $\mathcal{P}^{sbs}_d$ | The set of DL transmission powers for SBSs |
| $\mathcal{P}^{mbs}$ | The set of transmission powers for MBS to each SBS |
| $\Omega$ | The set of RA/BH transmission duration division |
| $\mathcal{D}$ | The set of radio frame configurations |
| $G_u, G_s, G_m, G_e$ | The non-cooperative games for SP1~SP4 |
| $\mathcal{U}, \mathcal{S}, \mathcal{M}, \mathcal{E}$ | The player sets of non-cooperative games $G_u, G_s, G_m, G_e$ |
| $\mu_u, \mu_s, \mu_m, \mu_e$ | The utility functions in non-cooperative games $G_u, G_s, G_m, G_e$ |
| $P^{ue}$ | The set of transmission powers that UEs have chosen |
| $P^{sbs}_u$ | The set of UL transmission powers that SBSs have chosen |
| $P^{sbs}_d$ | The set of DL transmission powers that SBSs have chosen |
| $P^{mbs}$ | The set of transmission powers that MBS has chosen for SBSs |
| $\Omega_{t,\tau}$ | The set of RA/BH transmission duration division that SBSs have chosen during $\tau$-th subframe of $t$-th frame |
| $D$ | The set of radio frame configurations that SBSs have chosen |

transmission duration division and UL/DL discrete power allocation for SBSs. The $SP3$ only focuses on the DL discrete power allocation optimization of the MBS to each SBS. The $SP4$ only concentrates on the optimization of RFCs of SBSs.

The $SP1$ is modeled as a non-cooperative game, where each UE acts as a game player to select its appropriate transmission power in order to maximize the system access throughput. The $SP2$ is modeled as a non-cooperative game, where each SBS acts as a game player to select its appropriate RA/BH transmission duration division and UL/DL transmission powers in order to maximize the system total throughput. The $SP3$ is modeled as a non-cooperative game, where the MBS is responsible for making game decisions for each SBS to select the appropriate transmission power from the MBS to each SBS in order to maximize the system backhaul throughput. The $SP4$ is modeled as a non-cooperative game, where the MBS is responsible for making game decisions for each SBS to select the appropriate RFC in order to maximize the system total throughput.

In order to get the suboptimal solution to the IOP, we propose the SL-HMF-RA scheme, which is a resource allocation scheme based on a single leader heterogeneous multiple followers Stackelberg game. In the SL-HMF-RA scheme, the game model of $SP4$ acts as an essential component of a single leader, which is deployed in the edge computing platform, while the game models of $SP1$, $SP2$, $SP3$ act as followers, which are deployed in each UE, each SBS, and the MBS, respectively.

The interaction process of all the participants is as follows: First, the leader initializes the SBSs' RFC set, the MBS's transmission power set, the SBSs' UL/DL transmission power set and transmission duration division set. Then, it broadcasts these initial results to each UE. The UEs follow a non-cooperative game to determine their own transmission power and then send the Nash equilibrium result of game to the leader. After receiving the results, the leader follows a non-cooperative game to determine RFCs, and then broadcasts the transmission power set of UEs, the transmission power set of MBS and RFCs to each SBS. After receiving the information, the SBSs follow a non-cooperative game to determine UL transmission power, DL transmission power, RA/BH transmission duration, and send the Nash equilibrium result of game to the leader. After receiving the results, the leader follows a non-cooperative game again to acquire RFCs and send the results receiving from SBSs, the transmission power of UEs and RFCs to the MBS. Similarly, the transmission power set of the MBS is also determined by a non-cooperative game after the MBS receives other players' information from leader. The MBS sends the determined transmission power set to the leader, and the leader determines optimized RFCs through a non-cooperative game again and broadcasts these results to each UE as at the beginning. The whole process described above repeats until the game reaches a Stackelberg Nash equilibrium.

In the following text, we will discuss the corresponding details about each non-cooperative game for each sub-problem and global Stackelberg game. Table 1 summarizes the relevant parameter variables of the IOP and the corresponding sub-problems.

### TABLE 1
### SUMMARY OF KEY NOTATIONS IN SECTION 5

#### 2) Problem Description and Solving Algorithm of SP1

When $P^{sbs}_u$, $P^{sbs}_d$, $P^{mbs}$, $D$, $\Omega_{t,\tau}$ are determined in advance, the **SP1** aims to find the set of transmission powers $P^{ue} = \{p_k^{ue} \in \mathcal{P}^{ue}\}_{k \in \mathcal{U}}$ for maximizing the system access throughput, which is expressed by

$$\begin{cases} SP1: \max_{\substack{PL^1, PL^2, PL^3 \\ \mathcal{P}^{ue}}} \left( T^{ura}_{PL^1} + T^{ura}_{PL^2} + T^{ura}_{PL^3} \right) \\ s.t. \ (40), (42a), (44), (47a), (49), (53a), C2, C4 \end{cases} \tag{72}$$

where the constraints unrelated to the $\mathcal{P}^{ue}$ are removed since $P^{sbs}_u$, $P^{sbs}_d$, $P^{mbs}$, $D$, $\Omega_{t,\tau}$ are fixed and satisfy relevant constraints.

When it comes to solving the **SP1**, the transmission power levels adopted by UE $u$ will affect the power levels of other UEs, and vice versa. The main reason is that other UEs may reuse the same frequency band and thus they interfere with each other. This situation forms a non-cooperative game process, which can be formulated by a potential game denoted by $G_u = [\mathcal{U}, \{p_k^{ue}\}_{k \in \mathcal{U}}, \{u_k\}_{k \in \mathcal{U}}]$, where the set of game players is $\mathcal{U} = \{1, 2, ..., U\}$. For a UE game player $k$, its strategy is $p_k^{ue} \in \mathcal{P}^{ue}$, and its utility function $u_k$ is a common function $\mu_u$. The common function $\mu_u$ is defined by

$$\mu_u = \begin{pmatrix} T^{ura}_{PL^1} + T^{ura}_{PL^2} \\ + T^{ura}_{PL^3} \end{pmatrix} + e_u \Phi_u \begin{pmatrix} T^{ura}_{PL^1}, T^{dbh}_{PL^1}, T^{ura}_{PL^2}, \\ T^{dbh}_{PL^2}, T^{ura}_{PL^3}, T^{dbh}_{PL^3} \end{pmatrix} \tag{73}$$

In (73), $e_u$ denotes the non-negative penalty scalar and its unit is "bps"; the function $\Phi_u()$ denotes the penalty function,

which is similar to the penalty function in [34]. $\Phi_u() = -1$ if the constraints (40), (42a), (44), (47a), (49), (53a), C2, and C4 are satisfied. $\Phi_u() = 0$ if these constraints are not satisfied. The first term in (73) corresponds to the part of the total utility related to the **SP1**, and the second term represents the constraints for throughput related to $\mathcal{P}^{ue}$, which imples a UE player that chooses a strategy violating these constraints will be punished.

For any UE $k$, the power allocation optimization process is shown in Algorithm 1, which can solve the **SP1**. Finally, the power allocation for all the UEs can be obtained. Notably, each UE $k$ generates the initial power $p_k^{ue}$ which is necessary in the beginning. The initial power $p_k^{ue}$ is the minimum available power from the set $\mathcal{P}^{ue}$, which ensures that the initial strategy profile is feasible under our previous assumption. In addition, the best response algorithm is used in the functional function $argmax_{\mu_u}()$, which is done by searching in the set $\mathcal{P}^{ue}$ in order from the minimum power value to the maximum one until the largest power value satisfying the constraints is obtained.

---

**Algorithm 1: UE Power Allocation Optimization Process**

Run at any UE $k$ ($k \in \mathcal{U}$) to act as the follower
**Input:** null
**Output:** $p^{ue}$

1: **If** receive the $(\mathcal{P}_d^{sbs}, \mathcal{P}_u^{sbs}, P^{mbs}, D, \Omega_{t,\tau})$ from the leader **then**
2:    Send the initial power of UE $k$ to the leader
3: **End if**
4: **If** receive the set $P^{ue}$ from the leader **then**
5:    $p_k^{ue} = argmax_{\mu_u}(\{p_k^{ue} \in P^{ue}\}_{k \in \mathcal{U} \setminus k}, P_u^{sbs}, P_d^{sbs}, P^{mbs}, D, \Omega_{t,\tau})$
6:    Send $p_k^{ue}$ to the leader
7: **End if**
8: **If** receive the $(P^{ue}, P_d^{sbs}, P_u^{sbs}, P^{mbs}, D, \Omega_{t,\tau})$ from the leader **then**
9:    $p_k^{ue} = argmax_{\mu_u}(\{p_k^{ue} \in P^{ue}\}_{k \in \mathcal{U} \setminus k}, P_d^{sbs}, P_u^{sbs}, P^{mbs}, D, \Omega_{t,\tau})$
10:    Send $p_k^{ue}$ to the leader
11: **End if**
12: **If** receive "end" from the leader **then** return **End if**

Run at the edge computing platform to act as a component of the leader
**Input:** the sets $P^{ue}, P_d^{sbs}, P_u^{sbs}, P^{mbs}, D, \Omega_{t,\tau}$
**Output:** null

1: Initialize the elements in the sets $P^{ue}$ and $\{E_k^{ue}\}_{k \in \mathcal{U}}$ to 0
2: **Repeat**
3:    **If** receive the power $p$ from any UE $k$ **then**
4:      **If** $p$ is equal to $p_k^{ue}$ ($p_k^{ue}$ is the element in the set $\mathcal{P}^{ue}$)
5:        $E_k^{ue} = 1$
6:      **Else**
7:        $p_k^{ue} = p$
8:      **End if**
9:      Broadcast the set $P^{ue}$ to all the UEs
10:    **End if**
11: **Until** all of the elements in the set $\{E_k^{ue}\}_{k \in \mathcal{U}}$ are 1
12: Send the "end" to all the UEs

---

### 3) Problem Description and Solving Algorithm of SP2

When $P^{ue}, P^{mbs}, D$ are determined in advance, the **SP2** aims to find the sets of transmission powers $\mathcal{P}_u^{sbs}$ and $\mathcal{P}_d^{sbs}$ and the set $\Omega_{t,\tau}$ for maximizing the system total throughput, which is expressed by

$$\begin{cases} SP2: \max_{\substack{P_{L1}, P_{L2}, P_{L3}, B_{PL2}^{mm}, P_{PL3}^{mm} \\ \mathcal{P}_u^{sbs}, \mathcal{P}_d^{sbs}, \Omega}} T_q \\ s.t. \quad (40),(42b),(44),(47b),(49),(53b),(55), \\ \qquad (57b),(59),(62b),(64),(68b),C1,C3,C4 \end{cases} \quad (74)$$

where the constraints unrelated to the $(\mathcal{P}_u^{sbs}, \mathcal{P}_d^{sbs}, \Omega)$ are removed since $P^{ue}, P^{mbs}, D$ are fixed and satisfy relevant constraints.

When it comes to solving the **SP2**, UL transmission power allocation, DL transmission power allocation, RA/BH transmission duration division adopted by SBS $m_m$ will affect decision of other SBSs, and vice versa. This situation also forms a non-cooperative game process, which can be formulated by a potential game denoted by $G_s = [\mathcal{S}, P_u^{sbs} \times \mathcal{P}_d^{sbs} \times D, \{s_s\}_{s \in \mathcal{M}^-}]$, where the set of game players is $\mathcal{S} = \mathcal{M}^-$. For a SBS game player $s$, its strategy set is $\mathcal{P}_u^{sbs} \times \mathcal{P}_d^{sbs} \times D$, and its utility function $s_s$ is a common function $\mu_s$. The common function $\mu_s$ is defined by

$$\mu_s = T_q + e_s \Phi_s \left( T_{PL1}^{ura}, T_{PL1}^{dbh}, T_{PL2}^{ura}, T_{PL2}^{dbh}, T_{PL3}^{ura}, T_{PL3}^{dbh} \right) \quad (75)$$

In (75), $e_s$ denotes the non-negative penalty scalar and its unit is "bps"; the function $\Phi_s()$ denotes the penalty function, which is similar to the penalty function in [34]. $\Phi_s() = -1$ if the constraints (40), (42b), (44), (47b), (49), (53b), (55), (57b), (59), (62b), (64), (68b), C1, C3, C4. $\Phi_s() = 0$ if these constraints are not satisfied. The first term in (75) corresponds to the part of the total utility related to the **SP2** and the second term represents the constraints for throughput related to $(\mathcal{P}_u^{sbs}, \mathcal{P}_d^{sbs}, \Omega)$, which imples any SBS player that chooses a strategy violating these constraints will be punished.

For any SBS $s$, the power allocation optimization and RA/BH transmission duration division are shown in Algorithm 2, which can solve the **SP2**. Notably, each SBS (e.g., $s$) selects the initial maximum UL transmission power from $\mathcal{P}_u^{sbs}$, the set of initial maximum DL transmission powers from $\mathcal{P}_d^{sbs}$, and the initial minimum RA/BH transmission division value from $\Omega$, which are necessary in the beginning to ensure that the initial strategy profile is feasible under our previous assumption. The sets $P^{ue}, P^{mbs}, D$ are also necessary which are considered to be fixed before Nash Equilibrium and sent by the leader run in edge computing platform. In addition, the best response algorithm is used in the functional function $argmax_{\mu_s}()$, which is done by searching in the sets $\mathcal{P}_u^{sbs}$ and $\mathcal{P}_d^{sbs}$ in order from the maximum value to the minimum one and the set $\Omega$ in order from the minimum value to the maximum one until the most desired values satisfying the constraints are obtained. The set $\mathcal{U}^s$ is the set communication peer ends of SBS $s$. The elements in set $\mathcal{U}^s$ are only UEs if SBS $s$ is a no-relay SBS, while they contain both UEs and another SBS.

---

**Algorithm 2: SBS Transmission Power Allocation and RA/BH Transmission Duration Division Optimization Process**

Run at any SBS $s$ (e.g., $s \in \mathcal{S}$) to act as the follower
**Input:** null
**Output:** $\{p_{d,s,k}^{sbs} \in \mathcal{P}_d^{sbs}\}_{k \in \mathcal{U}^s}, p_{u,s}^{sbs} \in \mathcal{P}_u^{sbs}, \gamma_{t,\tau}^s \in \Omega$

1: **If** receive the $(P^{ue}, P_d^{sbs}, P_u^{sbs}, P^{mbs}, D, \Omega_{t,\tau})$ from the leader **then**
2:    **If** SBS $s$ is a no-relay SBS **then**
3:      $\begin{Bmatrix} \{p_{d,s,k}^{sbs}\}_{k \in \mathcal{U}^s} \\ p_{u,s}^{sbs}, \gamma_{t,\tau}^s \end{Bmatrix} = argmax_{\mu_s} \left( \begin{matrix} \left( \{p_{d,s,k}^{sbs}\}_{k \in \mathcal{U}^s}, p_{u,s}^{sbs}, \gamma_{t,\tau}^s \right)_{s \in \mathcal{S} \setminus s}, \\ P^{ue}, P^{mbs}, D \end{matrix} \right)$
4:    **Else if** SBS $s$ is a relay SBS **then**
5:      $\begin{Bmatrix} \{p_{d,s,k}^{sbs}\}_{k \in \mathcal{U}^s} \\ p_{u,s}^{sbs} \end{Bmatrix} = argmax_{\mu_s} \left( \begin{matrix} \left( \{p_{d,s,k}^{sbs}\}_{k \in \mathcal{U}^s}, p_{u,s}^{sbs} \right)_{s \in \mathcal{S} \setminus s}, \\ P^{ue}, P^{mbs}, D, \Omega \end{matrix} \right)$
6:    **End if**

7: Send $\left\{\left\{p_{d,s,k}^{sbs}\right\}_{k\in\mathcal{U}^s}, p_{u,s}^{sbs}, \gamma_{t,\tau}^s\right\}$ to the leader

8: **End if**

9: **If** receive the $(P_d^{sbs}, P_u^{sbs}, \Omega_{t,\tau})$ from the leader **then**

10: **If** SBS $s$ is a no-relay SBS **then**

11: $\begin{Bmatrix}\left\{p_{d,s,k}^{sbs}\right\}_{k\in\mathcal{U}^s}, \\ p_{u,s}^s, \gamma_{t,\tau}^s\end{Bmatrix} = argmax_{\mu_s}\begin{pmatrix}\left\{p_{d,s,k}^{sbs}\right\}_{k\in\mathcal{U}^s}, p_{u,s}^{sbs}, \gamma_{t,\tau}^s \\ P^{ue}, P^{mbs}, D\end{pmatrix}$

12: **Else if** SBS $s$ is a relay SBS **then**

13: $\begin{Bmatrix}\left\{p_{d,s,k}^{sbs}\right\}_{k\in\mathcal{U}^s}, \\ p_{u,s}^s\end{Bmatrix} = argmax_{\mu_s}\begin{pmatrix}\left\{p_{d,s,k}^{sbs}\right\}_{k\in\mathcal{U}^s}, p_{u,s}^{sbs} \\ P^{ue}, P^{mbs}, D, \Omega\end{pmatrix}$

14: **End if**

15: Send $\left\{\left\{p_{d,s,k}^{sbs}\right\}_{k\in\mathcal{U}^s}, p_{u,s}^{sbs}, \gamma_{t,\tau}^s\right\}$ to the leader

16: **End if**

17: **If** receive "end" from the leader **then** return **End if**

Run at the edge computing platform to act as a component of the leader
**Input:** the sets $P^{ue}, P_d^{sbs}, P^{mbs}, D, \Omega_{t,\tau}$
**Output:** null

1: Initialize the elements in the sets $P_d^{sbs}, P_u^{sbs}, \Omega_{t,\tau}$ and $\{E_s^s\}_{s\in\mathcal{S}}$ to 0

2: **Repeat**

3: **If** receive $\{P_d, p_u, \gamma\}$ from any SBS $s$ **then**

4: **If** $\{P_d, p_u, \gamma\}$ is equal to $\left\{\left\{p_{d,s,k}^{sbs}\right\}_{k\in\mathcal{U}^s}, p_{u,s}^{sbs}, \gamma_{t,\tau}^s\right\}$

5: $E_s^s = 1$

6: **Else**

7: $\left\{\left\{p_{d,s,k}^{sbs}\right\}_{k\in\mathcal{U}^s}, p_{u,s}^{sbs}, \gamma_{t,\tau}^s\right\} = \{P_d, p_u, \gamma\}$

8: **End if**

9: Broadcast the $(P_d^{sbs}, P_u^{sbs}, \Omega_{t,\tau})$ to all the SBSs

10: **End if**

11: **Until** all the elements in the set $\{E_s^s\}_{s\in\mathcal{S}}$ are 1

12: Send the "end" to all the SBSs

### 4) Problem Description and Solving Algorithm of SP3

When $P^{ue}, P_u^{sbs}, P_d^{sbs}, D, \Omega_{t,\tau}$ are determined in advance, the **SP3** aims to find the set of transmission powers $P^{mbs} = \left\{p_{m_0,s}^{bs}\in\mathcal{P}^{mbs}\right\}_{s\in\mathcal{S}}$ for maximizing the system backhaul throughput, which is expressed by

$$SP3: \begin{cases} \max\limits_{P_L^1, P_L^2, P_L^3} \left(T_{PL^1}^{dbh} + T_{PL^2}^{dbh} + T_{PL^3}^{dbh}\right) \\ \mathcal{P}^{mbs} \\ s.t. \ (55), (57a), (59), (62a), (64), (68a), C4 \end{cases}$$ (76)

where the constraints unrelated to the $\mathcal{P}^{mbs}$ are removed since $P^{ue}, P_u^{sbs}, P_d^{sbs}, D, \Omega_{t,\tau}$ are fixed and satisfy relevant constraints.

Although only one MBS is involved in the paper, it has different transmission powers to different SBSs. Considering the scalability of SBSs, based on the idea of centralized resource allocation algorithm (CRA) [34], we model the **SP3** as a non-cooperative game. It can be formulated by a potential game denoted by $\bar{G}_m = [\mathcal{M}^-, \{p_{m_0,s}^{bs}\}_{s\in\mathcal{S}}, \{m_s\}_{s\in\mathcal{S}}]$, where we regard the decision of MBS for each power as a player in the game and the set of game players is $\mathcal{M}^- = \{m_1, ..., m_s, ..., m_M\}$. For a game player $s$, its strategy set is $P^{mbs}$, and its utility function $s_m$ is a common function $\mu_m$. The common function $\mu_m$ is defined by

$$\mu_m = \begin{pmatrix} T_{PL^1}^{dbh} + T_{PL^2}^{dbh} \\ + T_{PL^3}^{dbh} \end{pmatrix} + e_m\Phi_m\begin{pmatrix} T_{PL^1}^{ura}, T_{PL^1}^{dbh}, T_{PL^2}^{ura} \\ T_{PL^2}^{dbh}, T_{PL^3}^{ura}, T_{PL^3}^{dbh} \end{pmatrix}$$ (77)

In (77), $e_m$ denotes the non-negative penalty scalar and its unit is "bps"; the function $\Phi_m()$ denotes the penalty function, which is similar to the penalty function in [34]. $\Phi_m() = -1$ if the constraints (55), (57a), (59), (62a), (64), (68a), C4. $\Phi_m() = 0$ if these constraints are not satisfied. The first term in (77) corresponds to the part of the total utility related to the **SP3** and the second term represents the constraints for throughput related to $\mathcal{P}^{mbs}$, which imples the player that chooses a strategy violating these constraints will be punished.

The power allocation process for the MBS is shown in Algorithm 3, which can solve the **SP3**. Notably, the MBS selects the set of initial transmission powers from $\mathcal{P}^{mbs}$, which is necessary in the beginning. Each power in this set is used by the MBS to send data to a specific SBS. Each initial power is the minimum available power from $\mathcal{P}^{mbs}$, which ensures that the initial strategy profile is feasible under our previous assumption. The sets $P^{ue}, P_u^{sbs}, P_d^{sbs}, D, \Omega_{t,\tau}$ are also necessary, which are considered to be fixed before Nash Equilibrium and sent by the edge computing platform at which the leader runs. Compared to the **SP1** and **SP2**, the difference in solving **SP3** is that there is no need for the leader to collect and broadcast the players' decisions and no need for the leader to judge the end of the game.

**Algorithm 3: MBS Power Allocation Optimization Process**

Run at the MBS to act as the follower
**Input:** $P^{ue}, P_u^{sbs}, P_d^{sbs}, D, \Omega_{t,\tau}$
**Output:** $P^{mbs}$

1: **If** receive the $(P^{ue}, P_u^{sbs}, P_d^{sbs}, D, \Omega_{t,\tau})$ from the leader **then**

2: **Repeat**

3: **For** $\forall s \in \mathcal{M}^-$ **do**

4: $p_{m_0,m_s}^{bs} = argmax \ \mu_m\begin{pmatrix} P^{ue}, P_u^{sbs}, P_d^{sbs} \\ \{p_{m_0,m_s}^{bs}\}_{s\in\mathcal{S}\backslash s}, D, \Omega_{t,\tau} \end{pmatrix}$

5: **End for**

6: **Until** $\forall s \in \mathcal{M}^-$, the change of $\mu_m$ is less than a certain threshold

7: Send $P^{mbs}$ to the leader

8: **End if**

### 5) Problem Description and Solving Algorithm of SP4

When $P^{ue}, P_u^{sbs}, P_d^{sbs}, P^{mbs}, \Omega_{t,\tau}$ are determined in advance, the **SP4** aims to find the set $D$ for maximizing the system total throughput, which is expressed by

$$\begin{cases} \max\limits_{P_L^1, P_L^2, P_L^3} T_q \\ D \\ s.t. \ (40), (42), (44), (47), (49), (53), \\ (55), (57), (59), (62), (64), (68), C4 \end{cases}$$ (78)

where the constraints unrelated to the set $\mathcal{D}$ are removed since $P^{ue}, P_u^{sbs}, P_d^{sbs}, P^{mbs}, \Omega_{t,\tau}$ are fixed and satisfy relevant constraints.

Similarly to the **SP3**, the **SP4** is solved to obtain a set of RFCs for SBSs based on the idea of CRA algorithm in [34]. We model the **SP4** as a non-cooperative game, which is denoted by $G_e = [\mathcal{M}^-, \{d_s\}_{s\in\mathcal{S}}, \{e_s\}_{s\in\mathcal{S}}]$, where we regard the decision of edge computing platform for each SBS's RFC as a player in the game and the set of game players is $\mathcal{M}^-$. For a game player $s$, its strategy set is $D$, and its utility function $s_e$ is a common function $\mu_e$. The common function $\mu_e$ is defined by

$$\mu_e = T_q + e_e\Phi_e\begin{pmatrix} T_{PL^1}^{ura}, T_{PL^1}^{dbh}, T_{PL^1}^{ura} \\ T_{PL^2}^{dbh}, T_{PL^3}^{ura}, T_{PL^3}^{dbh} \end{pmatrix}$$ (79)

In (79), $e_e$ denotes the non-negative penalty scalar and their unit is "bps"; the function $\Phi_e()$ denotes the penalty function, which is similar to the penalty function in [34]. $\Phi_e() = -1$ if the constraints (40), (42), (44), (47), (49), (53), (55), (57), (59), (62), (64), (68), C4. $\Phi_e() = 0$ if these constraints are not satisfied. The first term in (79) corresponds to the part of the total utility related to the **SP4** and the second term represents the constraints for throughput related to the set $\mathcal{D}$, which implies the player that chooses a strategy violating these constraints will be punished. The RFC selection optimization process for SBSs is shown in Algorithm 4, which can solve the **SP4**.

---

**Algorithm 4: RFC Selection Optimization Process**

Run the edge computing platform to act as a component of the leader

**Input:** $P^{ue}, P_u^{sbs}, P_d^{sbs}, P^{mbs}, \Omega_{\tau,\tau}$

**Output:** $D, Q$

1: **Repeat**
2:    **For** $\forall s \in \mathcal{M}^-$ **do**
3:      $d_s = argmax\ \mu_e(P^{ue}, P_u^{sbs}, P_d^{sbs}, P^{mbs}, \{d_{s'}\}_{s' \in \mathcal{S}\backslash s}, \Omega_{\tau,\tau})$
4:    **End for**
5: **Until** $\forall s \in \mathcal{M}^-$, the change of $\mu_e$ is less than a certain threshold
6: $\mathcal{Q} = \mu_e(P^{ue}, P_u^{sbs}, P_d^{sbs}, P^{mbs}, \{d_{s'}\}_{s' \in \mathcal{S}\backslash s}, \Omega_{\tau,\tau})$

---

As aforesaid, the leader is deployed in the edge computing platform. The Stackelberg game-based SL-HMF-RA scheme is described in Algorithm 5. which can get an approximate solution to the **IOP**. To save space, the convergence proof of the above five algorithms is omitted. The interested readers can refer to Theorem 1 in [37] and Theorem 3 in [34] to verify the convergence of Algorithms 1~2 and Algorithms 3~4, respectively. Also, they can refer to Theorem 2 in [37] to verify the convergence of Algorithms 5.

---

**Algorithm 5: SL-HMF-RA Scheme**

This algorithm's leader runs at the edge computing platform

**Input:** $\varepsilon$ // $\varepsilon$ is a very small positive number

**Output:** $P^{ue}, P_u^{sbs}, P_d^{sbs}, P^{mbs}, D, \Omega_{\tau,\tau}$

1:   Initialize $P_u^{sbs}, P_d^{sbs}, D, \Omega_{\tau,\tau}$
2:   Broadcast $P_u^{sbs}, P_d^{sbs}, P^{mbs}, D, \Omega_{\tau,\tau}$ to all the UEs
3:   Invoke Algorithm 1 to get $P^{ue}$
4:   Invoke Algorithm 4 to get $D$ and $Q$
5:   Set iteration index $t = 0, Q_t = Q$
6:   Broadcast $P^{ue}, P_u^{sbs}, P_d^{sbs}, P^{mbs}, D, \Omega_{\tau,\tau}$ to all the SBSs
7:   Invoke Algorithm 2 to get $P_u^{sbs}, P_d^{sbs}, \Omega_{\tau,\tau}$
8:   Invoke Algorithm 4 to get $D$ and $Q$
9:   Update $t = t + 1, Q_t = Q$
10:   Send $P^{ue}, P_u^{sbs}, P_d^{sbs}, D, \Omega_{\tau,\tau}$ to all the MBS
11:   Invoke Algorithm 3 to get $P^{mbs}$
12:   Invoke Algorithm 4 to get $D$ and $Q$
13:   Update $t = t + 1, Q_t = Q$
14: **If** $|Q_t - Q_{t-1}| < \varepsilon$ **then**
15:    Send the "end" to all the UEs, SBSs and MBS
16: **Else**
17:    Broadcast $P^{ue}, P_u^{sbs}, P_d^{sbs}, P^{mbs}, D, \Omega_{\tau,\tau}$ to all the UEs
18:    Invoke Algorithm 1 to get $P^{ue}$
19:    Invoke Algorithm 4 to get $D$ and $Q$
20:    Update $t = t + 1, Q_t = Q$
21:    **If** $|Q_t - Q_{t-1}| < \varepsilon$ **then**
22:      Send the "end" to all the UEs, SBSs and MBS
23:    **Else**
24:      Go to 6
25:    **End if**
26: **End if**

---

# 6. PERFORMANCE EVALUATION

In this section, we will evaluate the performance of the proposed SL-HMF-RA scheme. The experimental parameter simulation settings and the simulation results are shown as follow. Every value shown in the figures of this section is the result of an average over 10 random instances.

*6.1 Experimental Parameter Simulation Settings*

Based on the cluster model in 3GPP [38] and the IAB simulation guidelines in 3GPP [3], we focus on a tree topology extracted from a 300m×300m service area. As shown in Fig. 3, one MBS is located at the tree root, three SBSs act as intermediate nodes, the other three SBSs act as leaf nodes, and six backhaul links make up the rest of the tree. Many UEs are randomly deployed in the service area. If not specified, the default number of UEs that are concurrently scheduled to access the network is 18. The considered heights of the MBS, SBSs and UEs are 25 m, 10 m and 1.5 m, respectively. We assume that BSs are equipped with 4 antenna panels, each of which has a RF chain and thus at most 4 concurrent connections associated with UEs. The main environment parameters can be found in TABLE 2.

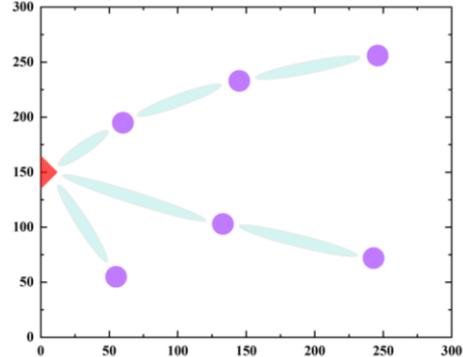

Fig. 3. The IAB network scenario considered in the experiments (red triangle: MBS, blue circles: SBSs, cyan ovals: backhaul links).

We assess the performance of our approach according to three levels of minimum spectral energy efficiency which the system should meet and refer to them as low spectral energy efficiency (LSEE), medium spectral energy efficiency (MSEE), and high spectral energy efficiency (HSEE). The $S_{EE}^{cons} = 0.32$ bits/J/Hz for LSEE, $S_{EE}^{cons} = 0.48$ bits/J/Hz for MSEE, and $S_{EE}^{cons} = 0.6$ bits/J/Hz for HSEE. The following experiments are performed under LSEE if not specifically emphasized.

TABLE 2
SIMULATION ENVIRONMENT PARAMETERS

| Symbol | Description | Value |
|---|---|---|
| $\epsilon$ | Side lobe gain | 0.001 |
| $\varphi^{mbs}$ | Beamwidth of the MBS | $10 \sim 60°$ |
| $\varphi^{sbs}$ | Beamwidth of the SBSs | $10 \sim 60°$ |
| $\varphi^{ue}$ | Beamwidth of the UEs | $5°$ |
| $b_r^{mm}$ | MmWave spectrum resource blocks | 1 GHz |
| $b_r^{thz}$ | THz spectrum resource blocks | 5 GHz |
| $p_{max}^{bs}$ | Maximum power of each BS | 44 dBm |
| $p_{max}^{ue}$ | Maximum power of each UE | 23 dBm |
| $L^{ue}$ | UE power division level | 20 |
| $L^{bs}$ | BS power division level | 10 |
| $Z$ | Number of time slots in each subframe | $10 \sim 30$ |
| $N_0$ | Background noise power spectrum density | -174 dBm/Hz |
| $f^{mm}$ | Carrier frequency for mmWave | 30 GHz |
| $f^{thz}$ | Carrier frequency for THz | 0.3 THz |
| $|\mathcal{M}^-|$ | Number of SBSs | 6 |
| $|\mathcal{U}|$ | Number of UEs | $18 \sim 36$ |
| $|\mathcal{B}_{PL^2}^{mm}|$ | Number of spectrum resource blocks used in Case 2 | 6 |
| $|\mathcal{B}_{PL^3}^{mm}|$ | Number of spectrum resource blocks used in Case 3 | 9 |

*6.2 Experimental Results and Analysis*

In our SL-HMF-RA scheme, according to the characteristics of the four sub-problems (i.e., SP1, SP2, SP3, SP4), the solutions to SP1 and SP2 are based on distributed game architecture, while those to SP3 and SP4 are based on centralized game architecture. For the convenience of the following description, our SL-HMF-RA scheme is renamed as *SL-HMF-RA-H*. In addition, the comparison schemes of this paper are described as follows.

● SL-HMF-RA with Centralized Resource Allocation Algorithm (referred to as *SL-HMF-RA-C*): The solutions to SP1, SP2, SP3, and SP4 are based on centralized game architecture.

● SL-HMF-RA with Genetic Algorithm (referred to as *SL-HMF-RA-G*): On the basis of *SL-HMF-RA-C* scheme, genetic algorithm is used to solve each sub-problem instead of non-cooperative game.

● SL-HMF-RA with Particle Swarm Optimization Algorithm (referred to as *SL-HMF-RA-P*): On the basis of SL-HMF-RA-G scheme, the genetic algorithm is replaced by the particle swarm optimization algorithm.

● SL-HMF-RA with Random Scheduling Algorithm (referred to as *SL-HMF-RA-R*): The SP1, SP2, SP3, and SP4 are solved by a random scheduling algorithm instead of a non-cooperative game, respectively.

**1) Convergence**

Fig. 4 shows the convergence of the *SL-HMF-RA-H* and *SL-HMF-RA-C* schemes in the same IAB network scenario. As the convergence behavior curves show very similar trends under LSEE, MSEE and HSEE conditions, we only show LSEE curves. The number of iterations is calculated in the schemes based on the value of the iteration variable. As we can see, compared with the *SL-HMF-RA-H* scheme, the *SL-HMF-RA-C* scheme reaches the convergence value in fewer iterations. This is because, in the *SL-HMF-RA-C* scheme, one iteration is a process in which the central node makes a decision for each player, while in the *SL-HMF-RA-H* scheme, one iteration is a process in which a participant makes a decision. It is worth noting that due to the different definitions of the iterations, fewer iterations do not mean less time to reach convergence.

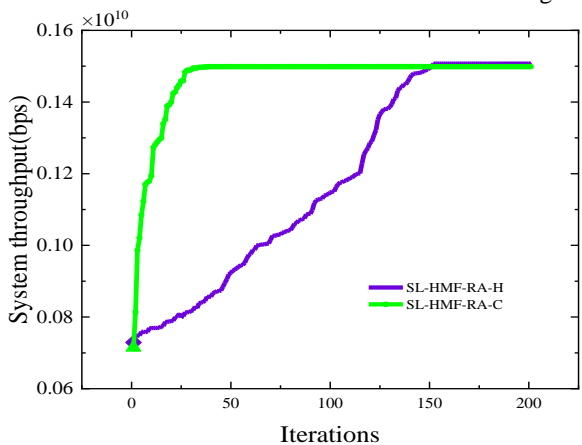

Fig. 4. Convergence of SL-HMF-RA-H and SL-HMF-RA-C.

**2) Throughput and Power Consumption in Different Schemes within Different Spectral Energy Efficiency**

Fig.5~Fig.7 indicate system throughput, UL throughput, and DL throughput in the LSEE, MSEE and HSEE scenarios,

respectively. Also, the corresponding power consumption values are shown in Fig.5~Fig.7.

From Fig. 5, it can be seen the system throughput for *SL-HMF-RA-H* and *SL-HMF-RA-C* schemes are similar under the three spectral energy efficiency conditions, and are significantly better than those of *SL-HMF-RA-G*, *SL-HMF-RA-P* and *SL-HMF-RA-R* schemes. The gap is mainly due to *SL-HMF-RA-H* and *SL-HMF-RA-C* schemes are constantly iterated through the non-cooperative game of each participant to obtain the suboptimal solutions to the subproblems, and *SL-HMF-RA-G*, *SL-HMF-RA-P* obtain the suboptimal solution of each dimension from the solution space according to the combinatorial optimization mathematical properties of the subproblems. These schemes perform best in the LSEE case, followed by the MSEE case, and finally the HSEE case. This is because the lower spectral energy efficiency constraint may lead to the higher transmission power, which is more beneficial to improve throughput. Moreover, it is evident that the order of the five schemes to total power consumption in LSEE case is *SL-HMF-RA-H*, *SL-HMF-RA-C*, *SL-HMF-RA-G*, *SL-HMF-RA-P*, *SL-HMF-RA-R*, while the order is reversed in MSEE and HSEE cases. This is because greater transmit power can be selected to improve throughput under the lower spectral energy efficiency constraint.

Fig. 6 and Fig. 7 provide an insight into throughput and power consumption for UL and DL. The order of the five schemes' throughput values is consistent with system throughput and also proves the above result analysis. The DL throughput is greater than the UL throughput, because the number of subframes with type D is greater than the number of subframes with type U in the RFC result. The DL power of SBSs is greater than the UL power of UEs in the access link, which is also the reason why the DL power consumption is higher than the UL power consumption. Furthermore, we can observe an interesting aspect that the UL and DL throughput comparisons of different schemes are consistent with the system throughput in the three spectral energy efficiency cases.

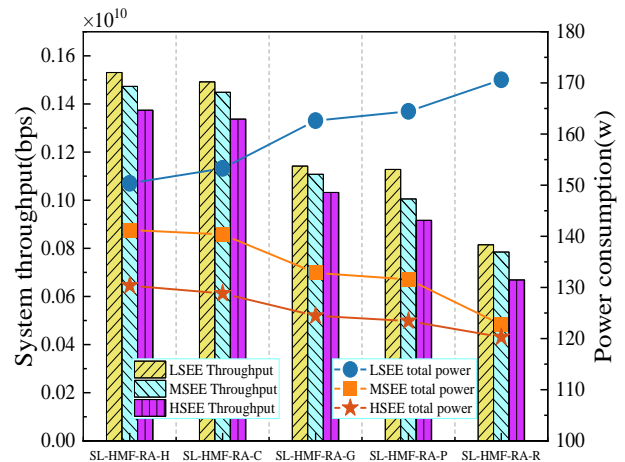

Fig. 5. System total throughput and total power consumption in different schemes within different spectral energy efficiency constraints.

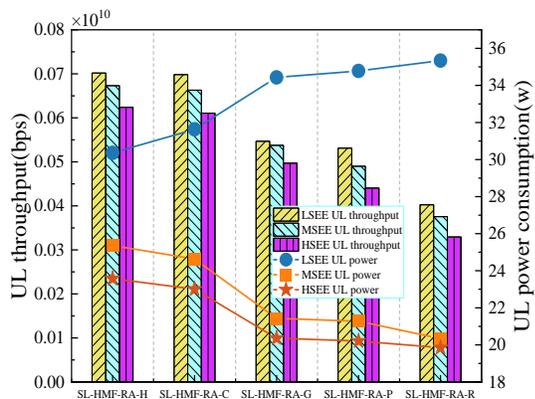

Fig. 6. UL throughput and UL power consumption in different schemes within different spectral energy efficiency constraints.

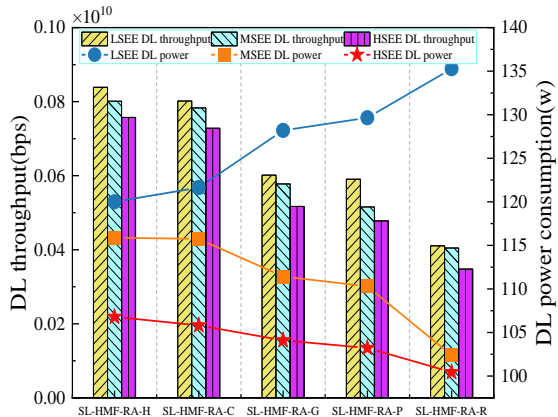

Fig. 7. DL throughput and DL power consumption in different schemes within different spectral energy efficiency constraints.

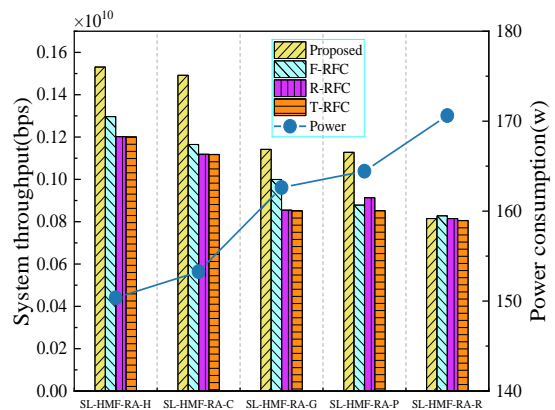

Fig. 8. System throughput and power consumption in different methods with different radio frame configurations.

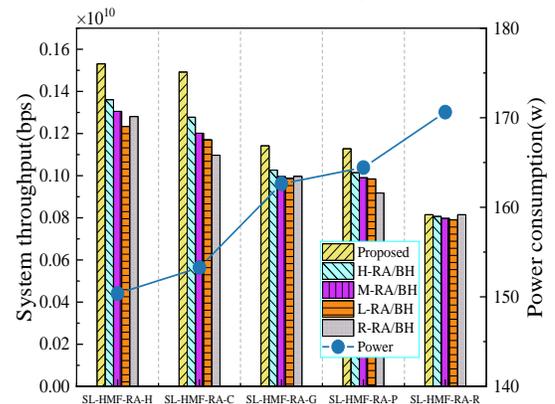

Fig. 9. System throughput and power consumption in different access and backhaul transmission duration division methods.

### 3) Throughput and Power Consumption under Different Methods for Specific Resources

It can be seen from Fig. 8 and Fig. 9, in order to explore management methods for specific resources and evaluate the throughput and power consumption of the schemes, we compare the proposed schemes with other schemes for radio frame configurations and RA/BH transmission duration allocation.

In Fig. 8, to evaluate the advantages of the proposed schemes for RFCs, we additionally considered the popular schemes for comprehensive comparisons, including: Fixed RFC (F-RFC) strategy, Random RFC (R-RFC) strategy, Traffic-matched RFC (T-RFC) strategy [11]. It can be observed that the proposed schemes outperform the comparison schemes for system throughput. This is attributed to the emphasis of the method in this paper on selecting reasonable RFCs. Besides, the power consumption of proposed scheme is the same as that of the corresponding comparison schemes, which means the changes to RFCs methods do not affect power consumption.

In Fig. 9, we considered High RA/BH transmission duration (H-RA/BH), Medium RA/BH transmission duration (M-RA/BH), Low RA/BH transmission duration (L-RA/BH), Random RA/BH transmission duration (R-RA/BH) comparison schemes, which are inferior to the proposed schemes. This is because that the method of this paper pays attention to optimizing RA/BH transmission duration. Similar to RFC, for RA/BH, the proposed schemes have the same power consumption as the corresponding comparison schemes.

### 4) Throughput and Power Consumption under Different Schemes

Fig. 10 compares the throughput and power consumption of the proposed schemes over different number of UEs. From the simulation results, we can find that the throughput of the above five schemes increases as the number of UEs increases. The increase in throughput is not proportional to the increase in the number of UEs, which is due to the greater interference associated with more access links, and the limited throughput. In the experimental results, the power consumption of the five schemes increases with the increase of UEs, which is also in line with common sense.

In Fig. 11, we compare the effect of the number of time slots in each subframe on the throughput and power consumption in the different schemes. It can be seen that no matter how many time slots in a scheduling period, the order of the five schemes is unchanged. With the increasing time slots in each scheduling period, the system throughput increases. Fig. 12 presents the effect of beamwidth of SBSs and UEs on the throughput and power consumption in the different schemes, where we assume all of SBSs and UEs have the same beamwidth for simplicity, and the beamwidth of MBS is set to 5°. As can be observed, with the increasing beamwidth, the throughput is gradually decreasing. This is mainly because a wider beamwidth is subject to potentially more interference sources. It can be observed from Fig. 11 and Fig. 12 that, both the number of time slots in the subframe and the beamwidth of SBSs and UEs have little effect on the power consumption.

Figs. 13 provides an insight into the average access throughput for UL and DL in the LSEE, MSEE and HSEE cases and the power consumption of associated UEs. The order of three spectral energy efficiency constraints and UL/DL to access throughput is consistent with system throughout. In addition, for the power consumption of UEs, it can be seen that the order of five schemes does not change under different spectral energy efficiency constraints. The *SL-HMF-RA-H* scheme is closed to the *SL-HMF-RA-C* scheme, but it has more power savings for UEs than the *SL-HMF-RA-G*, *SL-HMF-RA-P*, *SL-HMF-RA-R* schemes. The main reasons are the same as explained above.

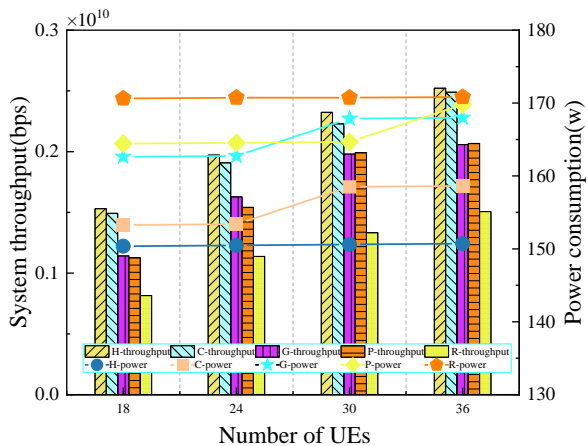

Fig. 10. System throughput and power consumption in different number of UEs.

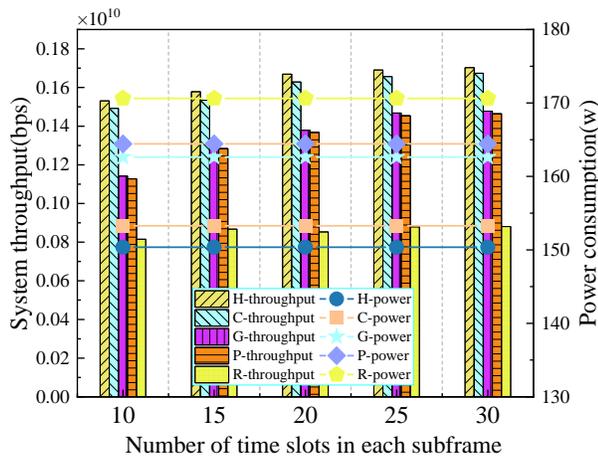

Fig. 11. System throughput and power consumption in different number of time slots in each subframe.

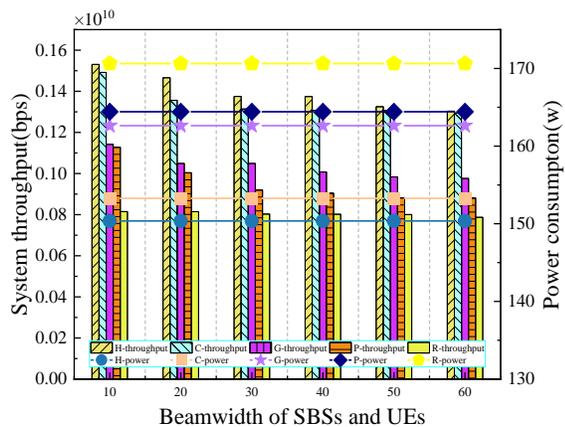

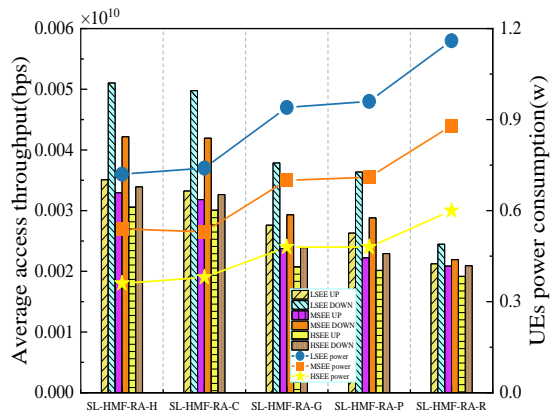

Fig. 12. System throughput and power consumption in different beamwidth of SBSs and UEs.

Fig. 13. Average access throughput for UL and DL and power consumption for UEs in different schemes within different spectral energy efficiency constraints.

# 7. CONCLUSIONS

In this paper, we have investigated the heterogeneous resource allocation problem in multi-hop IAB networks with D-TDD. We have addressed the integrated optimization problem of joint relaying SBS selection and UE association, discrete power control, non-unified RA/BH transmission duration allocation and radio frame configurations, in which the objective is to maximize the system throughput while meeting the end-to-end QoS constraints. Then, we have decomposed this integrated optimization problem with huge solution space into the several sub-problems with small solution space. These sub-problems can be solved easily by designing the methods based on non-cooperative games. Besides, Stackelberg game is applied to combine the solving results of all the sub-problems to get the approximate solution to the integrated optimization problem. The experimental simulation results have shown that our scheme can achieve good throughput performance.

CREDIT AUTHORSHIP CONTRIBUTION STATEMENT

DECLARATION OF COMPETING INTEREST

DATA AVAILABILITY

ACKNOWLEDGEMENT

# REFERENCES

[1] B. Yu, L. Yang, H. Ishii, and S. Mukherjee, Dynamic TDD support in macrocell-assisted small cell architecture, IEEE J. Sel. Areas Commun. 33 (6) (2015) 1201–1213.

[2] "NR; Integrated Access and Backhaul (IAB) radio transmission and reception," Technical Specification (TS) 38.174, Sept. 2024, v.18.6.0.

[3] Study on Integrated Access and Backhaul (Release 16), document 3GPP TR 38.874, V16.0.0, Jan. 2019.

[4] M. N. Kulkarni, J. G. Andrews, and A. Ghosh, Performance of dynamic and static TDD in self-backhauled millimeter wave cellular networks, IEEE Trans. Wireless Commun. 16 (10) (2017) 6460–6478.

[5] Interference Management in NR, document R1-1702719 TSG RAN WG1 Meeting #88, 3GPP, Sophia Antipolis, France, Feb. 2017.

[6] 3rd Generation Partnership Project; Technical Specification Group Radio Access Network; NR; Physical Layer Procedures for Control (Release 15) V15.5.0, 3GPP Standard TS 38.213, Mar. 2019.

[7] M. Kamel, W. Hamouda, and A. Youssef, Ultra-dense networks: A survey, IEEE Commun. Surveys Tuts. 18 (4) (2016) 2522–2545.

[8] V. D. Tuong, N. -N. Dao, W. Noh and S. Cho, Deep reinforcement learning-based hierarchical time division duplexing control for dense wireless and mobile networks, IEEE Trans. Wireless Commun. 20 (11) (2021) 7135–7150.

[9] J. -S. Tan et al., "Lightweight Machine Learning for Digital Cross-Link Interference Cancellation With RF Chain Characteristics in Flexible Duplex MIMO Systems," in IEEE Wireless Communications Letters, vol. 12, no. 7, pp. 1269-1273, July 2023.

[10] K. Boutiba, M. Bagaa and A. Ksentini, "Multi-Agent Deep Reinforcement Learning to Enable Dynamic TDD in a Multi-Cell Environment," in IEEE Transactions on Mobile Computing, vol. 23, no. 5, pp. 6163-6177, May 2024.

[11] K. Boutiba, M. Bagaa and A. Ksentini, "On enabling 5G Dynamic TDD by leveraging Deep Reinforcement Learning and O-RAN," NOMS 2023-2023 IEEE/IFIP Network Operations and Management Symposium, Miami, FL, USA, 2023.

[12] V. D. Tuong, W. Noh and S. Cho, "Spatial Deep Learning-Based Dynamic TDD Control for UAV-Assisted 6G Hotspot Networks," in IEEE Transactions on Industrial Informatics, vol. 20, no. 9, pp. 11092-11102, Sept. 2024.

[13] M. M. Razlighi, N. Zlatanov, S. R. Pokhrel, and P. Popovski, "Optimal centralized dynamic-time-division-duplex," IEEE Trans. Wireless Commun., vol. 20, no. 1, pp. 28–39, Jan. 2021.

[14] Q. Ding, C. Luo, H. Yang and Y. Luo, "Throughput Maximization of Dynamic TDD Networks With a Full-Duplex UAV-BS," in IEEE Transactions on Wireless Communications, vol. 23, no. 11, pp. 16821-16835, Nov. 2024.

[15] L. F. Abanto-Leon, A. Asadi, A. Garcia-Saavedra, G. H. Sim, and M. Hollick, "RadiOrchestra: Proactive management of millimeter-wave self-backhauled small cells via joint optimization of beamforming, user association, rate selection, and admission control," IEEE Trans. Wireless Commun., vol. 22, no. 1, pp. 153–173, Jan. 2023.

[16] Tang, Y. Zhou, and N. Kato, "Deep reinforcement learning for dynamic Uplink/Downlink resource allocation in high mobility 5G HetNet," IEEE J. Sel. Areas Commun., vol. 38, no. 12, pp. 2773–2782, Dec. 2020.

[17] M. Elsayed, M. Erol-Kantarci, and H. Yanikomeroglu, "Transfer reinforcement learning for 5G new radio mmWave networks," IEEE Trans. Wireless Commun., vol. 20, no. 5, pp. 2838 – 2849, May 2021.

[18] Guo, L. Tang, X. Zhang, and Y.-C. Liang, "Joint optimization of handover control and power allocation based on multi-agent deep reinforcement learning," IEEE Trans. Veh. Technol., vol. 69, no. 11, pp. 13124–13138, Nov. 2020.

[19] S. Shen, Y. Ren, Y. Ju, X. Wang, W. Wang and V. C. M. Leung, "EdgeMatrix: A Resource-Redefined Scheduling Framework for SLA-Guaranteed Multi-Tier Edge-Cloud Computing Systems," in IEEE Journal on Selected Areas in Communications, vol. 41, no. 3, pp. 820-834, March 2023, doi: 10.1109/JSAC.2022.3229444.

[20] B. Zhang, F. Devoti, I. Filippini, and D. De Donno, "Resource allocation in mmWave 5G IAB networks: A reinforcement learning approach based on column generation," Comput. Netw., vol. 196, Sep. 2021, Art. no. 108248.

[21] Q. Cheng, Z. Wei, and J. Yuan, "Deep reinforcement learning-based spectrum allocation and power management for IAB networks," in Proc. IEEE Int. Conf. Commun. Workshops (ICC Workshops), Jun. 2021, pp. 1–6.

[22] M. Pagin, T. Zugno, M. Polese, and M. Zorzi, "Resource management for 5G NR integrated access and backhaul: A semi-centralized approach," IEEE Trans. Wireless Commun., vol. 21, no. 2, pp. 753–767, Feb. 2022.

[23] C. Huang and X. Wang, "A Bayesian Approach to the Design of Backhauling Topology for 5G IAB Networks," in IEEE Transactions on Mobile Computing, vol. 22, no. 4, pp. 1867-1879, 1 April 2023.

[24] S. Gopalam, S. V. Hanly and P. Whiting, "Distributed Resource Allocation and Flow Control Algorithms for mmWave IAB Networks," in IEEE/ACM Transactions on Networking, vol. 31, no. 6, pp. 3175-3190, Dec. 2023.

[25] M. Sheng, Y. Zhang, J. Liu, Z. Xie, T. Q. S. Quek and J. Li, "Enabling Integrated Access and Backhaul in Dynamic Aerial-Terrestrial Networks for Coverage Enhancement," in IEEE Transactions on Wireless Communications, vol. 23, no. 8, pp. 9072-9084, Aug. 2024.

[26] B. Yu et al., "Realizing High Power Millimeter Wave Full Duplex: Practical Design, Prototype and Results," GLOBECOM 2022 - 2022 IEEE Global Communications Conference, Rio de Janeiro, Brazil, 2022, pp. 3803-3808.

[27] A. A. Gargari, A. Ortiz, M. Pagin, W. de Sombre, M. Zorzi and A. Asadi, "Risk-Averse Learning for Reliable mmWave Self-Backhauling," in IEEE/ACM Transactions on Networking, vol. 32, no. 6, pp. 4989-5003, Dec. 2024.

[28] Q. Xue, X. Fang, C. X. Wang, Beamspace SU-MIMO for future millimeter wave wireless communications, IEEE J. Sel. Areas Commun. 35 (7) (2017) 1564–1575.

[29] T. Bai, R. W. Heath, Coverage and rate analysis for millimeterwave cellular networks, IEEE Trans. Wireless Commun. 14 (2) (2015) 1100–1114.

[30] V. Petrov, M. Komarov, D. Moltchanov, J.M. Jornet, Y. Koucheryavy, Interference and SINR in millimeter wave and terahertz communication systems with blocking and directional antennas, IEEE Transactions on Wireless Communications 16 (3) (2017) 1791–1808.

[31] J. M. Jornet and I. F. Akyildiz, Channel modeling and capacity analysis for electromagnetic wireless nanonetworks in the terahertz band, IEEE Trans. Wireless Commun. 10 (10) (2011) 3211–3221.

[32] L. S. Rothman et al. HITRAN: High-resolution transmission molecular absorption database harvard-smithson center for astrophysics (2014). [Online]. Available: https://www.cfa.harvard.edu.

[33] NR and NG-RAN Overall Description, Standard 3GPP TS 38.300, 3GPP, 2017.

[34] Y. P. Liu, X. M. Fang, and M. Xiao, Discrete power control and transmission duration allocation for self-backhauling dense mmWave cellular networks, IEEE Transactions on Communications 66 (1) (2018) 432–447.

[35] N. Naderializadeh, J. J. Sydir, M. Simsek, H. Nikopour, Resource management in wireless networks via multi-agent deep reinforcement learning, IEEE Transactions on Wireless Communications 20 (6) (2021) 3507–3523.

[36] P. Belotti, C. Kirches, S. Leyffer, J. Linderoth, J. Luedtke, and A. Mahajan, Mixed-integer nonlinear optimization, Acta Numerica 22 (2013) 1–131.

[37] J. S. Gui, L. Yin, X. H. Deng, L. Cai, Joint access and relay-assisted backhaul resource allocation for dense mmWave multiple access networks, IEEE Transactions on Vehicular Technology 73 (1) (2024) 1289–1305.

[38] Small cell enhancements for E-UTRA and E-UTRAN—Physical layer aspects, document 3GPP TR 36.872, V.1.0.0, Rel. 12, Aug. 2013.